\documentclass[preprint,12pt]{elsarticle}

\usepackage{amssymb}
\usepackage{xcolor}
\usepackage{amsmath, amssymb}
\usepackage{booktabs}
\usepackage{geometry}
\usepackage{graphicx}
\usepackage{multirow}
\usepackage{adjustbox}
\usepackage{float} 
\usepackage{subcaption} 

\begin{document}

\begin{frontmatter}



\title{A Unified Framework for EEG Seizure Detection Using Universum-Integrated Generalized Eigenvalues Proximal Support Vector Machine}



\author[inst2]{Yogesh Kumar}
\ead{yogesh.23csz0014@iitrpr.ac.in}

\author[inst2]{Vrushank Ahire}
\ead{2022csb1002@iitrpr.ac.in}

\author[inst2]{M. A. Ganaie\corref{cor1}}
\ead{mudasir@iitrpr.ac.in}
\cortext[cor1]{Corresponding author}

\affiliation[inst2]{organization={Department of Computer Science and Engineering, Indian Institute of Technology Ropar},
            city={Rupnagar},
            postcode={140001}, 
            state={Punjab},
            country={India}}

\begin{abstract}

The paper presents novel Universum-enhanced classifiers: the Universum Generalized Eigenvalue Proximal Support Vector Machine (U-GEPSVM) and the Improved U-GEPSVM (IU-GEPSVM) for EEG signal classification. Using the computational efficiency of generalized eigenvalue decomposition and the generalization benefits of Universum learning, the proposed models address critical challenges in EEG analysis: non-stationarity, low signal-to-noise ratio, and limited labeled data. U-GEPSVM extends the GEPSVM framework by incorporating Universum constraints through a ratio-based objective function, while IU-GEPSVM enhances stability through a weighted difference-based formulation that provides independent control over class separation and Universum alignment. The models are evaluated on the Bonn University EEG dataset across two binary classification tasks: (O~vs~S)—healthy (eyes closed) vs seizure, and (Z~vs~S)—healthy (eyes open) vs seizure. IU-GEPSVM achieves peak accuracies of 85\% (O~vs~S) and 80\% (Z~vs~S), with mean accuracies of 81.29\% and 77.57\% respectively, outperforming baseline methods.
Rigorous statistical validation confirms these improvements: Friedman tests reveal significant overall differences, pairwise Wilcoxon signed-rank tests with Bonferroni correction establish IU-GEPSVM's superiority over all baselines, and win-tie-loss analysis demonstrates practical significance. Overall, integrating interictal Universum data yields an efficient and reliable solution for neurological diagnosis.
\end{abstract}

\end{frontmatter}



\section{Introduction}

Electroencephalogram (EEG) signal classification has become an important area of research in computational neuroscience and machine learning, with significant applications in brain-computer interfaces (BCIs) \cite{lotte2018review} and the diagnosis of neurological disorders \cite{ganaie2022brain}. The non-stationary nature of EEG signals, along with their low signal-to-noise ratio, presents unique challenges that have spurred the development of advanced machine learning techniques. Traditional methods, such as Support Vector Machines (SVMs) \cite{guido2024overview}, have proven effective in EEG classification tasks, particularly for detecting epileptic seizures \cite{richhariya2018eeg} and monitoring cognitive states \cite{amin2017classification}. However, the increasing complexity of EEG analysis requires more sophisticated solutions that can effectively manage noise, high dimensionality, and limited labeled data while ensuring computational efficiency.

Incorporating prior knowledge about EEG data distributions, such as interictal signals or resting-state activity, can significantly improve the performance of classifiers. This understanding has led to the integration of Universum learning—a framework that uses ``non-class" samples to enhance decision boundaries-into modern Support Vector Machine (SVM) models. Recent research by \cite{ganaie2023eeg} and \cite{quadir2024intuitionisticE} has shown that Universum-based methods outperform traditional SVMs in noisy EEG environments, especially when combined with robust loss functions like pinball loss or intuitionistic fuzzy membership.

The evolution of EEG classification methodologies has evolved in several parallel paths. Early approaches primarily focused on feature extraction techniques combined with conventional classifiers. For example, \cite{amin2017classification} demonstrated impressive accuracy by using wavelet-based features in conjunction with SVM classifiers. Similarly, \cite{zhang2015magnetic} achieved high classification accuracy by integrating the stationary wavelet transform with generalized eigenvalue proximal SVM (GEPSVM).  These studies highlighted the significance of appropriate feature selection in EEG analysis, with wavelet transforms \cite{zhang2015magnetic, zhang2015preclinical} and entropy measures \cite{zhang2015preclinical} proving particularly effective for capturing the temporal and spectral characteristics of brain signals. In recent years, researchers have started improving EEG classification not only by focusing on better features but also by making the models more robust and capable of learning from extra background data, known as Universum data \cite{ma2025adaptive,hazarika2024eeg}

A significant advancement in the efficiency of Support Vector Machines (SVMs) came with the development of Twin Support Vector Machines (TWSVMs) \cite{khemchandani2007twin}. This approach reduced computational complexity by solving two smaller quadratic programming problems (QPPs) instead of one large QPP. Subsequent innovations, such as Universum TWSVM (UTSVM) \cite{qi2012twin} and Granular Ball TWSVM using universum data (GBU-TSVM) \cite{ganaie2025granulara}, further enhanced the robustness of the model against outliers and noise—issues that are particularly challenging in EEG analysis. Some of the latest methods, like UTPMSVM \cite{hazarika2024eeg} and Pin-UTSVM \cite{ganaie2023eeg}, use Universum data and special loss functions to make EEG classification more stable and noise resistant. In the same direction, FULSTSVM \cite{richhariya2022fuzzy} and GBU-TSVM \cite{ganaie2025granulara}  include fuzzy and granular ideas to better handle noisy and uncertain data. IFUTSVM-ID \cite{tanveer2024robust} goes one step further by dealing effectively with imbalanced EEG data, which often occurs in real-world medical situations.

The introduction of proximal support vector machines via generalized eigenvalues (GEPSVM), as presented by \cite{mangasarian2005multisurface}, marked a significant advancement in classification efficiency. This approach solves two generalized eigenvalue problems to generate non-parallel hyperplanes, providing computational advantages over traditional support vector machines (SVMs) while maintaining competitive accuracy. Subsequent improvements to GEPSVM included the development of improved GEPSVM (IGEPSVM) \cite{shao2012improved}, which addressed singularity issues using standard eigenvalue decomposition, and manifold regularized GEPSVM \cite{liang2016manifold}, which incorporated local geometric information. The versatility of GEPSVM variants was further demonstrated in multi-view learning scenarios \cite{sun2018multiview} and in medical image classification \cite{zhang2015magnetic}, highlighting their potential for electroencephalogram (EEG) analysis. Recent versions of GEPSVM, such as $L_{2,p}$-GEPSVM \cite{yan2022robust}, have been designed to deal with different kinds of outliers and still train quickly. Similarly, ULSTELM \cite{wu2025universum} combines Universum data with an extreme learning machine framework to achieve better accuracy even with fewer labeled EEG samples.

Parallel to these developments, the concept of Universum learning emerged as a powerful paradigm for incorporating prior knowledge into classification tasks. The use of Universum data, which refers to samples that do not belong to any target class but capture meaningful variations, shows potential to improve generalization \cite{chapelle2007analysis}. This concept was later adapted for EEG classification through various SVM extensions. Richhariya and Tanveer \cite{richhariya2018eeg} proposed a Universum SVM (USVM) for EEG signal classification, while \cite{qi2012twin} developed a Universum twin SVM (UTSVM) that achieved superior performance by placing Universum data in a nonparallel insensitive loss tube. Recent work has further enhanced these approaches through regularization techniques \cite{gupta2019regularized}, intuitionistic fuzzy methods \cite{quadir2024intuitionisticE}, and pinball loss functions \cite{ganaie2023eeg} to improve robustness against noise and outliers. In addition to these, AdaBoost-based SQSSVM \cite{ma2025adaptive} has been proposed to make classification faster and stronger by combining multiple weak models. Pin-MvUTSVM \cite{lou2024multi} adds multi-view learning and pinball loss to improve stability across different EEG feature sets. TPMSVM \cite{peng2011tpmsvm} provides an efficient parametric-margin framework, which forms the base for several later Universum-based classifiers.

Current methods for analyzing EEG signals, despite recent advancements, face two main limitations: (1) sensitivity to label inaccuracies, and (2) insufficient incorporation of prior knowledge about EEG signal distributions. Our proposed methodology builds upon the GEPSVM framework \cite{mangasarian2005multisurface} while incorporating Universum constraints similar to those in UTSVM \cite{qi2012twin}, with additional robustness enhancements inspired by recent work on intuitionistic fuzzy methods \cite{quadir2024intuitionisticE} and pinball loss \cite{ganaie2023eeg}. The algorithm maintains GEPSVM's computational efficiency through generalized eigenvalue decomposition while gaining the generalization benefits of Universum learning, particularly valuable for EEG applications where interictal or resting-state data can serve as effective Universum samples \cite{richhariya2018eeg, kumar2021universum}.

The motivation behind this work is to address the persistent challenges in EEG seizure detection caused by the noisy and non-stationary characteristics of EEG signals, along with the limited availability of labeled clinical data. Existing classifiers often overlook domain-specific prior information, such as interictal EEG segments, which can improve model generalization and diagnostic reliability. Universum learning offers a systematic way to incorporate such non-class samples, enhancing decision boundaries and reducing sensitivity to noise and outliers. However, this concept has not been effectively integrated with computationally efficient eigenvalue-based classifiers. Therefore, combining Universum learning with the GEPSVM framework forms the central motivation of this work, aiming to develop a stable and efficient approach capable of managing uncertainty and improving the clinical effectiveness of EEG-based seizure detection.

Key contributions:
\begin{itemize}
    \item The proposed Universum-Integrated Generalized Eigenvalue Proximal Support Vector Machine (U-GEPSVM) embeds Universum learning directly within the generalized eigenvalue formulation, enhancing decision boundary generalization in EEG seizure classification.
    \item An Improved U-GEPSVM (IU-GEPSVM) introduces a weighted difference-based objective that mitigates numerical instability and provides independent control over class separation and Universum alignment.
    \item An efficient Universum selection strategy is incorporated, leveraging interictal EEG segments and wavelet-based representations, supported by  Principal Component Analysis (PCA) \cite{abdi2010principal}, and Independent Component Analysis (ICA) \cite{comon1994independent} for feature extraction and dimensionality reduction.
    \item Experimental evaluation on the Bonn University EEG dataset demonstrates that the proposed models achieve notable improvements over GEPSVM, I-GEPSVM, and UTSVM in both classification accuracy and robustness.
    \item The study establishes a unified, computationally stable framework that integrates domain-aware Universum learning with eigenvalue-based classifiers, offering clinically relevant advancements for EEG-based seizure detection.
\end{itemize}

The remainder of this paper is organized to provide thorough coverage of both theoretical foundations and practical applications. Section 2 reviews related work in greater depth, Section 3 presents the mathematical formulation of U-GEPSVM, Section 4 describes experimental methodology and results, and Section 5 concludes with discussion and future directions. This structure ensures readers gain an understanding of the method's development, implementation, and performance characteristics.

\section{Related Work}

This section provides an overview of the foundational concepts and methodologies relevant to our research, including Twin Support Vector Machine (TWSVM), Universum Twin Support Vector Machine (U-TWSVM), Generalized Eigenvalue Proximal SVM (GEPSVM), and Improved Generalized Eigenvalue Proximal SVM (IGEPSVM).

\subsection{Preliminaries}
Here we introduce the mathematical foundations and notation used throughout the paper, focusing on generalized eigenvalue problems and the Rayleigh quotient, which are central to the discussed methodologies.

\subsubsection{Notation and Definitions}
\begin{itemize}
    \item[$\scalebox{0.6}{$\blacksquare$}$] Let $X_1 \in \mathbb{R}^{m_1 \times n}$ and $X_2 \in \mathbb{R}^{m_2 \times n}$ denote the matrices representing data points for Class 1 and Class 2, respectively, where $m_1 + m_2 = m$ and $n$ denotes the number of features.
    \item[$\scalebox{0.6}{$\blacksquare$}$] The Universum data, which provides additional contextual information, is represented by $U \in \mathbb{R}^{p \times n}$.
    \item[$\scalebox{0.6}{$\blacksquare$}$] The weight vectors and bias terms for the two hyperplanes are $w_1, w_2 \in \mathbb{R}^n$ and $b_1, b_2 \in \mathbb{R}$, respectively.
    \item[$\scalebox{0.6}{$\blacksquare$}$] Vectors of ones for each class are denoted as $e_1 \in \mathbb{R}^{m_1}$, $e_2 \in \mathbb{R}^{m_2}$, and $e_U \in \mathbb{R}^p$.
    \item[$\scalebox{0.6}{$\blacksquare$}$] Regularization parameters are $\delta \in \mathbb{R}$, $\gamma_1, \gamma_2 \in \mathbb{R}$ for class weighting, and $\Psi_1, \Psi_2 \in \mathbb{R}$ for Universum weighting.
\end{itemize}

\subsubsection{Generalized Eigenvalue Problem}
The generalized eigenvalue problem \cite{ghojogh2019eigenvalue} is a fundamental concept in linear algebra that extends the standard eigenvalue problem to include two matrices. It involves finding values $\lambda$ and non-zero vectors z that satisfy:
\begin{equation}
    G z = \lambda H z, \quad z \neq 0,
\end{equation}
where $G$ and $H$ are symmetric matrices of size $(n+1) \times (n+1)$. This formulation has important properties:

\begin{itemize}
    \item[$\scalebox{0.8}{$\blacktriangleright$}$] If $H$ is positive definite, the eigenvalues $\lambda$ are real and finite, making computation stable.
    \item[$\scalebox{0.8}{$\blacktriangleright$}$] The eigenvectors $z$ identify directions where the transformation by $G$ is proportional to the transformation by $H$.
\end{itemize}

This mathematical framework serves as the foundation for many machine learning algorithms, particularly in dimensionality reduction and classification tasks. For instance, Linear Discriminant Analysis (LDA) utilizes this framework where $G$ and $H$ represent between-class and within-class scatter matrices, respectively.

\subsubsection{Rayleigh Quotient}
The Rayleigh quotient \cite{parlett1974rayleigh} provides a measure of how a matrix scales a vector in a specific direction. For a symmetric matrix $G$, the Rayleigh quotient is defined as:

\begin{equation}
    r(z) = \frac{z^T G z}{z^T z}
\end{equation}
In the context of generalized eigenvalue problems, this extends to:
\begin{equation}
    r(z) = \frac{z^T G z}{z^T H z},
\end{equation}
where $H$ is positive definite. The significance of the Rayleigh quotient lies in its extremal properties: it attains its minimum and maximum values at the eigenvectors corresponding to the smallest and largest eigenvalues of $G z = \lambda H z$. This property makes it an essential tool for optimization in various machine learning models, as it provides a way to find directions that maximize or minimize specific criteria.

\subsection{Twin Support Vector Machine (TWSVM)}

TWSVM \cite{khemchandani2007twin} is an advancement over traditional Support Vector Machines (SVMs) by constructing two non-parallel hyperplanes instead of a single one. Each hyperplane is specifically designed to be close to one class while maintaining a significant distance from the other class. The key innovation is that each hyperplane focuses primarily on its designated class: \( f^+ \) is constructed to be close to Class 1 and far from Class 2, while \( f^- \) serves the opposite purpose.

Given the data matrices \( X_1 \) (representing Class 1) and \( X_2 \) (representing Class 2), TWSVM defines the two hyperplanes as follows:
\begin{equation}
    f^+(x) = w_1^T x + b_1 = 0, \quad f^-(x) = w_2^T x + b_2 = 0.
\end{equation}

To find the optimal hyperplanes, TWSVM solves two separate quadratic programming problems:

Optimization problem for the positive class:
\begin{equation}
\begin{aligned}
    \min_{w_1, b_1, \xi} \quad & \frac{1}{2} \|X_1 w_1 + e_1 b_1\|^2 + c_1 e_2^T \xi \\
    \text{s.t.} \quad 
    & -(X_2 w_1 + e_2 b_1) + \xi \geq e_2, \\
    & \xi \geq 0.
\end{aligned}
\end{equation}

Optimization problem for the negative class:

\begin{equation}
\begin{aligned}
    \min_{w_2, b_2, \eta} \quad & \frac{1}{2} \|X_2 w_2 + e_2 b_2\|^2 + c_2 e_1^T \eta \\
    \text{s.t.} \quad 
    & (X_1 w_2 + e_1 b_2) + \eta \geq e_1, \\
    & \eta \geq 0.
\end{aligned}
\end{equation}
In these formulations, \( c_1 \) and \( c_2 \) are regularization parameters that control the trade-off between model complexity and training errors. The slack variables \( \xi \) and \( \eta \) allow for soft margins, permitting some data points to violate the constraints while incurring a penalty.

By applying Lagrangian duality, the optimization problems can be transformed into their dual forms:
\begin{equation}
    \max_{\alpha} e_2^T \alpha - \frac{1}{2} \alpha^T G (H^T H)^{-1} G^T \alpha \quad \text{s.t.} \quad 0 \leq \alpha \leq c_1 e_2,
\end{equation}

\begin{equation}
    \max_{\beta} e_1^T \beta - \frac{1}{2} \beta^T P (Q^T Q)^{-1} P^T \beta \quad \text{s.t.} \quad 0 \leq \beta \leq c_2 e_1,
\end{equation}

where $H = [X_1 \ e_1]$, $G = [X_2 \ e_2]$, $P = [X_1 \ e_1]$, and $Q = [X_2 \ e_2]$.

Once these dual problems are solved, the parameters of the hyperplanes can be calculated as:
\begin{equation}
    z_1 = \begin{bmatrix} w_1 \\ b_1 \end{bmatrix} = -(H^T H)^{-1} G^T \alpha, \quad z_2 = \begin{bmatrix} w_2 \\ b_2 \end{bmatrix} = -(Q^T Q)^{-1} P^T \beta.
\end{equation}
The advantage of TWSVM over conventional SVM lies in its ability to capture more complex decision boundaries by using two hyperplanes, which can lead to improved classification performance, especially in the case of non-linearly separable data.

\subsection{Universum Twin Support Vector Machine (UTSVM)}

UTSVM  \cite{qi2012twin} extends the TWSVM framework by incorporating additional knowledge in the form of Universum data. The Universum concept refers to a set of examples that do not belong to either class but provide contextual information about the problem domain. By incorporating this information, UTSVM aims to improve generalization performance.

The inclusion of Universum data $U$ leads to modified optimization problems:

Optimization problem for the positive class:
\begin{equation}
\begin{aligned}
    \min_{w_1, b_1, \xi, \psi} \quad & \frac{1}{2} \|X_1 w_1 + e_1 b_1\|^2 + c_1 e_2^T \xi + c_u e_U^T \psi \\
    \text{s.t.} \quad 
    & -(X_2 w_1 + e_2 b_1) + \xi \geq e_2, \\
    & (U w_1 + e_U b_1) + \psi \geq (-1 + \epsilon) e_U, \\
    & \xi \geq 0,\quad \psi \geq 0.
\end{aligned}
\end{equation}

Optimization problem for the negative class:
\begin{equation}
\begin{aligned}
    \min_{w_2, b_2, \eta, \psi^*} \quad & \frac{1}{2} \|X_2 w_2 + e_2 b_2\|^2 + c_2 e_1^T \eta + c_u e_U^T \psi^* \\
    \text{s.t.} \quad 
    & (X_1 w_2 + e_1 b_2) + \eta \geq e_1, \\
    & -(U w_2 + e_U b_2) + \psi^* \geq (-1 + \epsilon) e_U, \\
    & \eta \geq 0,\quad \psi^* \geq 0.
\end{aligned}
\end{equation}

In these formulations, $c_u$ is a regularization parameter that controls the influence of Universum data, and $\epsilon$ is a parameter that determines how close the Universum data should be to the hyperplanes. The new constraints involving Universum data encourage the hyperplanes to pass through or close to the Universum examples, helping to refine the decision boundary in regions where class discrimination is most challenging.

The dual problems now include additional Lagrange multipliers ($\mu$ and $\nu$) corresponding to the Universum constraints:
\begin{equation}
    \max_{\alpha, \mu} -\frac{1}{2} (\alpha^T G - \mu^T O)(H^T H)^{-1} (G^T \alpha - O^T \mu) + e_2^T \alpha + (\epsilon - 1) e_U^T \mu
\end{equation}
\begin{equation}
    \text{s.t.} \quad 0 \leq \alpha \leq c_1 e_2, \quad 0 \leq \mu \leq c_u e_U.
\end{equation}
\begin{equation}
    \max_{\lambda, \nu} -\frac{1}{2} (\lambda^T P - \nu^T S)(Q^T Q)^{-1} (P^T \lambda - S^T \nu) + e_1^T \lambda + (\epsilon - 1) e_U^T \nu
\end{equation}
\begin{equation}
    \text{s.t.} \quad 0 \leq \lambda \leq c_2 e_1, \quad 0 \leq \nu \leq c_u e_U.
\end{equation}
Here, the matrices $O = [U \ e_U]$ and $S = [U \ e_U]$ incorporate the Universum data into the optimization framework. The integration of Universum data represents a significant enhancement to TWSVM by leveraging additional knowledge to guide the formation of decision boundaries, particularly in areas where class separation is ambiguous.

\subsection{Generalized Eigenvalue Proximal SVM (GEPSVM)}
GEPSVM \cite{mangasarian2005multisurface} takes a different approach to binary classification by framing it as a generalized eigenvalue problem rather than a quadratic programming problem. This approach focuses on finding two non-parallel hyperplanes by minimizing the distance of each class to its corresponding hyperplane while maximizing the distance to the other hyperplane.

GEPSVM defines two non-parallel hyperplanes:

\begin{equation}
    x^T w_1 + b_1 = 0, \quad x^T w_2 + b_2 = 0.
\end{equation}

The core idea of GEPSVM is to find hyperplanes that minimize the ratio of squared distances:

First Hyperplane:

\begin{equation}
    \min_{(w_1, b_1) \neq 0} \frac{\|X_1 w_1 + e_1 b_1\|^2 + \delta \|z_1\|^2}{\|X_2 w_1 + e_2 b_1\|^2}, \quad z_1 = \begin{bmatrix} w_1 \\ b_1 \end{bmatrix}.
\end{equation}

This ratio represents the squared distance of Class 1 points to the first hyperplane (plus a regularization term) divided by the squared distance of Class 2 points to the same hyperplane. The parameter $\delta$ is a regularization term that helps prevent overfitting and ensures numerical stability.

This optimization problem can be reformulated as a generalized eigenvalue problem:
\begin{equation}
    G z_1 = \lambda H z_1, \quad G = [X_1 \ e_1]^T [X_1 \ e_1] + \delta I, \quad H = [X_2 \ e_2]^T [X_2 \ e_2].
\end{equation}
Second Hyperplane:
\begin{equation}
    \min_{(w_2, b_2) \neq 0} \frac{\|X_2 w_2 + e_2 b_2\|^2 + \delta \|z_2\|^2}{\|X_1 w_2 + e_1 b_2\|^2}, \quad z_2 = \begin{bmatrix} w_2 \\ b_2 \end{bmatrix}.
\end{equation}
This leads to another generalized eigenvalue problem:
\begin{equation}
    L z_2 = \lambda M z_2, \quad L = [X_2 \ e_2]^T [X_2 \ e_2] + \delta I, \quad M = [X_1 \ e_1]^T [X_1 \ e_1].
\end{equation}
The solution to each generalized eigenvalue problem is the eigenvector corresponding to the smallest eigenvalue. These eigenvectors directly yield the hyperplane parameters $(w_1, b_1)$ and $(w_2, b_2)$.

The advantage of GEPSVM is that it avoids the complex quadratic programming of traditional SVMs and provides a more geometric interpretation of classification. However, it can face challenges when dealing with ill-conditioned matrices or when the denominator in the ratio becomes very small.

\subsection{Improved Generalized Eigenvalue Proximal SVM (I-GEPSVM)}

I-GEPSVM, proposed by \cite{shao2012improved}, addresses several limitations of GEPSVM by reformulating the optimization problem and introducing a weighting parameter. This approach replaces generalized eigenvalue problems with standard eigenvalue problems, introduces a weighting parameter to balance class-specific distances and avoids potential singularity issues present in GEPSVM.

Given data matrices $X_1 \in \mathbb{R}^{m_1 \times n}$ (Class 1) and $X_2 \in \mathbb{R}^{m_2 \times n}$ (Class 2), IGEPSVM seeks two non-parallel hyperplanes:  
\begin{equation}
    w_1^T x + b_1 = 0 \quad \text{and} \quad w_2^T x + b_2 = 0,
\end{equation}  
Unlike GEPSVM, which uses a ratio-based objective, IGEPSVM employs a difference measure between the distances:

First Hyperplane Optimization:
\begin{equation}
    \min_{(w_1, b_1) \neq 0} \frac{\|X_1 w_1 + e_1 b_1\|^2}{\|w_1\|^2 + b_1^2} - \nu \frac{\|X_2 w_1 + e_2 b_1\|^2}{\|w_1\|^2 + b_1^2},
\end{equation}  
Here, $\nu > 0$ is a crucial parameter that adjusts the trade-off between the proximity to Class 1 and the distance from Class 2. This formulation aims to find a hyperplane that is close to Class 1 points while being adequately distant from Class 2 points.

Second Hyperplane Optimization:
\begin{equation}
    \min_{(w_2, b_2) \neq 0} \frac{\|X_2 w_2 + e_2 b_2\|^2}{\|w_2\|^2 + b_2^2} - \nu \frac{\|X_1 w_2 + e_1 b_2\|^2}{\|w_2\|^2 + b_2^2}.
\end{equation}  
With Tikhonov regularization (parameter $\delta > 0$) and defining $z_i = [w_i; b_i]$, the optimization problems reduce to:  
\begin{equation}
    \min_{z_1} z_1^T (M - \nu H) z_1 + \delta \|z_1\|^2
\end{equation}
and
\begin{equation}
    \min_{z_2} z_2^T (H - \nu M) z_2 + \delta \|z_2\|^2,
\end{equation}  
where $M = [X_1 \ e_1]^T [X_1 \ e_1]$ and $H = [X_2 \ e_2]^T [X_2 \ e_2]$. 

These can be solved as standard eigenvalue problems:  
\begin{equation}
    (M + \delta I - \nu H) z_1 = \lambda_1 z_1 \quad \text{and} \quad (H + \delta I - \nu M) z_2 = \lambda_2 z_2.
\end{equation}  
After obtaining the two hyperplanes, a new data point $x$ is classified according to the shortest perpendicular distance to either hyperplane:  
\begin{equation}
    \text{Class}(x) = \arg\min_{i=1,2} \frac{|w_i^T x + b_i|}{\|w_i\|}.
\end{equation}  
The introduction of IGEPSVM represents a significant advancement in non-parallel hyperplane classification methods, offering both theoretical elegance and practical improvements in classification performance.

\section{Proposed U-GEPSVM and IU-GEPSVM Models}

This section presents our novel Universum-enhanced GEPSVM formulations that take advantage of both labeled training data and additional Universum samples to improve classification performance. The proposed models extend the GEPSVM framework in two distinct ways: U-GEPSVM maintains the ratio-based objective while incorporating Universum data, whereas IU-GEPSVM introduces a more stable difference-based formulation with Universum integration.

\subsection{Geometric Interpretation of U-GEPSVM}

Figure 1 illustrates the geometric behavior of U-GEPSVM in a two-dimensional feature space, demonstrating how two non-parallel hyperplanes partition the space while incorporating Universum constraints.
\begin{figure}[H]
\centering
\includegraphics[width=0.6\textwidth]{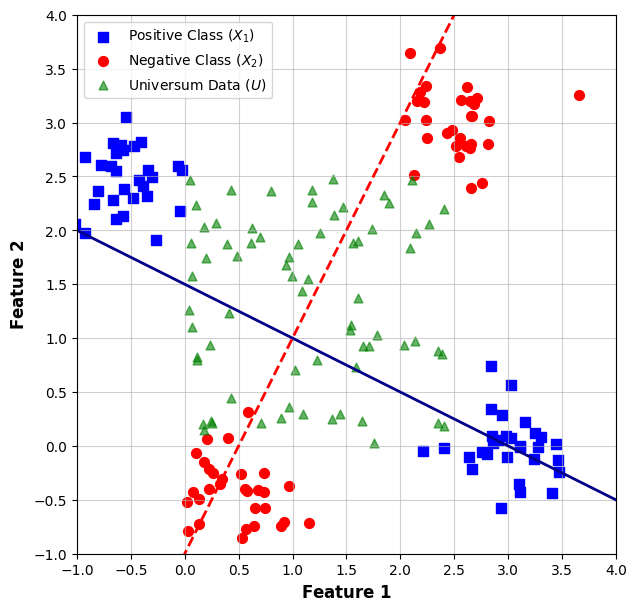}
\caption{Geometric interpretation of U-GEPSVM showing two non-parallel hyperplanes separating positive class (blue squares), negative class (red circles), with Universum data (green triangles) refining the decision boundary in ambiguous regions.}
\label{fig:geometric_interpretation}
\end{figure}
The positive class ($X_1$, blue squares) concentrates in opposing diagonal quadrants, while the negative class ($X_2$, red circles) occupies complementary regions. Critically, Universum samples (U, green triangles) populate the intermediate zones where class boundaries overlap. The solid blue hyperplane minimizes distance to $X_1$ while maximizing distance from both $X_2$ and U (Eq. \eqref{EQ28}), and conversely for the dashed red hyperplane (Eq. \eqref{EQ40}).
Unlike labeled classes, Universum data appear only in the denominator of the objective function (Equations \eqref{EQ31} , \eqref{EQ52}), creating a ``repulsion effect" that pushes hyperplanes away from ambiguous regions. For EEG seizure detection, interictal signals naturally fulfill this role, representing transitional brain states that share properties with both healthy and ictal conditions but belong to neither category. This geometric configuration directly contributes to the 19-28\% accuracy improvements demonstrated in Section 4, where explicitly modeling intermediate physiological states enhances generalization over methods ignoring such domain knowledge.

\subsection{U-GEPSVM: Universum Generalized Eigenvalue Proximal SVM}
Here we discuss the linear and non-linear optimization problems associated with the proposed U-GEPSVM.  

\subsubsection{Linear U-GEPSVM}

The U-GEPSVM model seeks two non-parallel hyperplanes where each plane is proximal to samples of one class, distal from samples of the opposite class, and properly aligned with Universum samples that represent ``non-examples" of both classes. For class-specific data matrices $X_1 \in \mathbb{R}^{m_1 \times n}$ (Class +1) and $X_2 \in \mathbb{R}^{m_2 \times n}$ (Class -1), and Universum data $U \in \mathbb{R}^{p \times n}$, we optimize for the first plane:

\begin{equation}
\label{EQ28}
\min_{(w_1,b_1)\neq 0} \frac{\|X_1w_1 + e_1b_1\|^2 + \delta\|z_1\|^2}{\|X_2w_1 + e_2b_1\|^2 + \|Uw_1 + e_ub_1\|^2}
\end{equation}

where $z_1 = [w_1, b_1]^T$, $e_1$, $e_2$, and $e_u$ are vectors of ones with appropriate dimensions, and $\delta > 0$ is a Tikhonov regularization parameter. The numerator of this objective ensures that the hyperplane is close to Class 1 samples, while the denominator ensures it is far from both Class 2 and Universum samples. The regularization term $\delta\|z_1\|^2$ helps prevent overfitting and ensures numerical stability.

To solve this optimization problem, we redefine it using the following matrices:

\begin{equation}
G = [X_1 \; e_1]^T[X_1 \; e_1] + \delta I,\; H = [X_2 \; e_2]^T[X_2 \; e_2]
\end{equation}

\begin{equation}
P = [U \; e_u]^T[U \; e_u],\; z_1 = \begin{bmatrix} w_1 \\ b_1 \end{bmatrix}
\end{equation}

The optimization problem can be rewritten as:

\begin{equation}
\label{EQ31}
\min_{z_1 \neq 0} \frac{z_1^T G z_1}{z_1^T H z_1 + z_1^T P z_1}
\end{equation}

To solve this ratio-based optimization problem, we introduce the constraint $z_1^T (H+P) z_1 = \alpha_1$ for some $\alpha_1 > 0$. This allows us to formulate the Lagrangian:
\vspace{-2mm}
\begin{align}
\label{eq32} \nonumber
\mathcal{L}(z_1, \lambda_1, \lambda_1^*) &= z_1^T G z_1 - \lambda_1 \left( z_1^T (H + P) z_1 - \alpha_1 \right) - \lambda_1^* \alpha_1 \\ 
&= z_1^T G z_1 - \lambda_1 z_1^T (H + P) z_1 + (\lambda_1 - \lambda_1^*) \alpha_1
\end{align}
\vspace{-0.5mm}
where $\lambda_1$ and $\lambda_1^*$ are Lagrange multipliers. We need to compute the gradient of this Lagrangian with respect to $z_1$. For the term $z_1^T G z_1$, we can express it as a double summation:
\begin{equation}
z_1^T G z_1 = \sum_{i=1}^{n+1}\sum_{j=1}^{n+1} z_{1i} g_{ij}  z_{1j}
\end{equation}
The partial derivative with respect to a particular component $z_{1k}$ is:
\begin{equation}
\frac{\partial}{\partial z_{1k}}\left(\sum_{i=1}^{n+1}\sum_{j=1}^{n+1} z_{1i} g_{ij}  z_{1j}\right) = \sum_{j=1}^{n+1} g_{kj} z_{1j} + \sum_{i=1}^{n+1} z_{1i} g_{ik} 
\end{equation}
Since $G$ is symmetric (i.e., $g_{ij} = g_{ji}$), these two sums are equal, yielding:
\begin{equation}
\frac{\partial}{\partial z_{1k}}(z_1^T G z_1) = 2\sum_{j=1}^{n+1} g_{kj} z_{1j} = 2[Gz_1]_k
\end{equation}
In vector form, the gradient becomes:

\begin{equation}
\frac{\partial}{\partial z_1}(z_1^T G z_1) = 2Gz_1
\end{equation}
Similarly, the gradients of the other quadratic terms are:
\begin{equation}
\frac{\partial}{\partial z_1}(z_1^T H z_1) = 2Hz_1 \text{ and } \frac{\partial}{\partial z_1}(z_1^T P z_1) = 2Pz_1
\end{equation}
Taking the full gradient of the Lagrangian (see Equation \eqref{eq32}) with respect to $z_1$ and setting it to zero:
\begin{align}
\nabla_{z_1} \mathcal{L} &= 2G z_1 - 2\lambda_1 (H + P) z_1 = 0 \\
\label{eq39}
\Rightarrow G z_1 &= \lambda_1 (H + P) z_1
\end{align}
The solution $z_1 = [w_1, b_1]^T$ corresponds to the eigenvector associated with the smallest eigenvalue $\lambda_1$ of this system. This eigenvector provides the optimal hyperplane parameters that balance proximity to Class 1 samples and distance from both Class 2 and Universum samples.

Similarly, for the second plane, we solve:

\begin{equation}
\label{EQ40}
\min_{(w_2,b_2)\neq 0} \frac{\|X_2w_2 + e_2b_2\|^2 + \delta\|z_2\|^2}{\|X_1w_2 + e_1b_2\|^2 + \|Uw_2 + e_ub_2\|^2}
\end{equation}

Using matrices defined as:
\begin{equation}
M = [X_2 \; e_2]^T[X_2 \; e_2] + \delta I,\; N = [X_1 \; e_1]^T[X_1 \; e_1]
\end{equation}
\begin{equation}
Q = [U \; e_u]^T[U \; e_u],\; z_2 = \begin{bmatrix} w_2 \\ b_2 \end{bmatrix}
\end{equation}
Following the same procedure of introducing constraints, formulating the Lagrangian, and taking gradients, we arrive at the generalized eigenvalue problem:
\begin{equation}
\label{eq43}
Mz_2 = \lambda_2(N+Q)z_2
\end{equation}
The solution $z_2 = [w_2, b_2]^T$ is the eigenvector corresponding to the smallest eigenvalue $\lambda_2$. The transformation $Gz_1$ ensures that the hyperplane $z_1$ is as close as possible to class 1 data points, while $\lambda_1(H+P)z_1$ ensures that the hyperplane $z_1$ is as far as possible from class 2 and Universum data points.

\subsubsection{Kernel U-GEPSVM}

For nonlinear decision boundaries, we employ the kernel trick by mapping data to a high-dimensional feature space $\mathcal{H}$ via a mapping function $\phi(\cdot)$. For the first hyperplane, the kernelized objective becomes:
\begin{equation}
\min_{(w_1,b_1)\neq 0} \frac{\|\Phi(X_1)w_1 + e_1b_1\|^2 + \delta\|z_1\|^2}{\|\Phi(X_2)w_1 + e_2b_1\|^2 + \|\Phi(U)w_1 + e_ub_1\|^2}
\end{equation}
where $\Phi(X)$ represents the matrix of mapped feature vectors. To solve this optimization problem efficiently without explicitly computing the high-dimensional mapping, we use the representer theorem, which states that the solution $w_1$ can be expressed as a linear combination of the mapped data points:
\begin{equation}
w_1 = \sum_{i=1}^{m_1} \alpha_i \phi(x_i^1) + \sum_{j=1}^{m_2} \beta_j \phi(x_j^2) + \sum_{k=1}^p \gamma_k \phi(u_k)
\end{equation}
where $x_i^1$, $x_j^2$, and $u_k$ are data points from Class 1, Class 2, and the Universum set, respectively.

Let us define the combined data matrix $Z = [X_1; X_2; U]$ and the corresponding mapped matrix $\Phi(Z)$. Then, $w_1 = \Phi(Z)^T \alpha$ for some coefficient vector $\alpha$. Substituting this into our objective function and using the kernel function $K(x_i, x_j) = \phi(x_i)^T \phi(x_j)$, we can reformulate our objective.

For the numerator term $\|\Phi(X_1)w_1 + e_1b_1\|^2$, we have:
\begin{align}
\|\Phi(X_1)w_1 + e_1b_1\|^2 &= \|\Phi(X_1)\Phi(Z)^T \alpha + e_1b_1\|^2\\
&= \|K_{1Z}\alpha + e_1b_1\|^2
\end{align}
where $K_{1Z}$ is the kernel matrix between $X_1$ and $Z$.
Similarly, for the denominator terms:
\begin{align}
\|\Phi(X_2)w_1 + e_2b_1\|^2 &= \|K_{2Z}\alpha + e_2b_1\|^2\\
\|\Phi(U)w_1 + e_ub_1\|^2 &= \|K_{UZ}\alpha + e_ub_1\|^2
\end{align}
The regularization term becomes:
\begin{equation}
\|z_1\|^2 = \|w_1\|^2 + b_1^2 = \alpha^T K_{ZZ} \alpha + b_1^2
\end{equation}
Defining the augmented vectors and matrices:
\begin{equation}
\tilde{K}_{1Z} = [K_{1Z} \; e_1], \tilde{K}_{2Z} = [K_{2Z} \; e_2], \tilde{K}_{UZ} = [K_{UZ} \; e_u], \tilde{\alpha} = [\alpha; b_1]
\end{equation}
The optimization problem is given as:
\begin{equation}
\label{EQ52}
\min_{\tilde{\alpha} \neq 0} \frac{\|\tilde{K}_{1Z}\tilde{\alpha}\|^2 + \delta\|\tilde{\alpha}\|^2}{\|\tilde{K}_{2Z}\tilde{\alpha}\|^2 + \|\tilde{K}_{UZ}\tilde{\alpha}\|^2}
\end{equation}
Defining matrices:
\begin{equation}
G_K = \tilde{K}_{1Z}^T \tilde{K}_{1Z} + \delta I, H_K = \tilde{K}_{2Z}^T \tilde{K}_{2Z}, P_K = \tilde{K}_{UZ}^T \tilde{K}_{UZ}
\end{equation}
We arrive at a generalized eigenvalue problem similar to the linear case:
\begin{equation}
G_K \tilde{\alpha} = \lambda_1 (H_K + P_K) \tilde{\alpha}
\end{equation}
To solve this, we introduce the constraint $\tilde{\alpha}^T (H_K + P_K) \tilde{\alpha} = \gamma$ for some $\gamma > 0$ and formulate the Lagrangian:
\begin{equation}
L(\tilde{\alpha}, \lambda_1) = \tilde{\alpha}^T G_K \tilde{\alpha} - \lambda_1(\tilde{\alpha}^T (H_K + P_K) \tilde{\alpha} - \gamma) - \lambda_1^*\gamma
\end{equation}
Computing the gradient with respect to $\tilde{\alpha}$ involves the double summation form:
\begin{equation}
\tilde{\alpha}^T G_K \tilde{\alpha} = \sum_{i=1}^{m}\sum_{j=1}^{m} \tilde{\alpha}_i 
 (G_K)_{ij} \tilde{\alpha}_j
\end{equation}
Taking the derivative with respect to $\tilde{\alpha}_k$:
\begin{equation}
\frac{\partial}{\partial \tilde{\alpha}_k}\left(\sum_{i=1}^{m}\sum_{j=1}^{m} (G_K)_{ij} \tilde{\alpha}_i \tilde{\alpha}_j\right) = 2\sum_{j=1}^{m} (G_K)_{kj} \tilde{\alpha}_j = 2[G_K\tilde{\alpha}]_k
\end{equation}
Setting the complete gradient to zero:
\begin{equation}
2G_K\tilde{\alpha} - 2\lambda_1(H_K + P_K)\tilde{\alpha} = 0
\end{equation}
This gives us the generalized eigenvalue problem:
\begin{equation}
G_K\tilde{\alpha} = \lambda_1(H_K + P_K)\tilde{\alpha}
\end{equation}
The eigenvector $\tilde{\alpha}$ corresponding to the smallest eigenvalue gives us the solution. Similarly, for the second hyperplane, we formulate and solve an analogous generalized eigenvalue problem in kernel space.

For classifying a new data point $x$, we compute the perpendicular distances to both hyperplanes in the feature space and assign the point to the class whose hyperplane is closer:
\begin{equation}
\text{class}(x) = \arg\min_{i=1,2} \frac{|\sum_{j} K(z_j, x) \alpha_j^i + b_i|}{\sqrt{\sum_{j,k} \alpha_j^i K(z_j, z_k) \alpha_k^i }}
\end{equation}
where $\alpha^i$ are the coefficients for the $i$-th hyperplane.

\subsection{IU-GEPSVM: Improved Universum GEPSVM}
Here we discuss the linear and non-linear optimization problems associated with the proposed IU-GEPSVM.  

\subsubsection{Linear IU-GEPSVM}
Unlike the ratio-based GEPSVM, our IU-GEPSVM employs a weighted difference formulation that avoids numerical instability when denominators are small and provides separate control over class separation and Universum alignment. For the first plane, the objective function is:
\begin{equation}
\min_{(w_1,b_1)\neq 0} \left(\frac{\|X_1w_1 + e_1b_1\|^2}{\|w_1\|^2 + b_1^2} - \gamma_1\frac{\|X_2w_1 + e_2b_1\|^2}{\|w_1\|^2 + b_1^2} - \psi_1\frac{\|Uw_1 + e_ub_1\|^2}{\|w_1\|^2 + b_1^2}\right)
\end{equation}

where $\gamma_1$ controls the emphasis on separating Class 1 from Class 2 and $\psi_1$ governs the influence of Universum samples. Using the matrix notation:
\begin{equation}
G = [X_1 \; e_1]^T[X_1 \; e_1],\; H = [X_2 \; e_2]^T[X_2 \; e_2]
\end{equation}
\begin{equation}
P = [U \; e_u]^T[U \; e_u],\; z_1 = \begin{bmatrix} w_1 \\ b_1 \end{bmatrix}
\end{equation}
The objective function can be simplified as:
\begin{equation}
\min_{z_1 \neq 0} \frac{z_1^T G z_1}{z_1^T z_1} - \gamma_1\frac{z_1^T H z_1}{z_1^T z_1} - \psi_1\frac{z_1^T P z_1}{z_1^T z_1}
\end{equation}
To solve this, we introduce the constraint $z_1^T z_1 = \alpha_1$ for some $\alpha_1 > 0$ and add Tikhonov regularization:
\begin{equation}
\min_{z_1,\alpha_1} \frac{1}{\alpha_1}z_1^T G z_1 + \delta\|z_1\|^2 - \frac{\gamma_1}{\alpha_1}z_1^T H z_1 - \frac{\psi_1}{\alpha_1}z_1^T P z_1
\end{equation}
\begin{equation}
\text{s.t. } z_1^T z_1 = \alpha_1, \alpha_1 > 0
\end{equation}
We formulate the Lagrangian:
\begin{equation}
\mathcal{L}(z_1, \lambda_1) = z_1^T G z_1 + \delta\|z_1\|^2 - \gamma_1 z_1^T H z_1 - \psi_1 z_1^T P z_1 - \lambda_1(z_1^T z_1 - \alpha_1) - \lambda_1^*\alpha_1
\end{equation}
where $\lambda_1$ and $\lambda_1^*$ are Lagrange multipliers. To find the critical points, we need to compute the gradient with respect to $z_1$. For each quadratic term, we use the double summation representation. For instance:
\begin{equation}
z_1^T G z_1 = \sum_{i=1}^{n+1}\sum_{j=1}^{n+1} z_{1i} g_{ij} z_{1j}
\end{equation}
The partial derivative with respect to component $z_{1k}$ is:
\begin{equation}
\frac{\partial}{\partial z_{1k}}\left(\sum_{i=1}^{n+1}\sum_{j=1}^{n+1} z_{1i} g_{ij} z_{1j}\right) = \sum_{j=1}^{n+1} g_{kj} z_{1j} + \sum_{i=1}^{n+1} g_{ik} z_{1i}
\end{equation}
Since $G$ is symmetric, this simplifies to:
\begin{equation}
\frac{\partial}{\partial z_{1k}}(z_1^T G z_1) = 2\sum_{j=1}^{n+1} g_{kj} z_{1j} = 2[Gz_1]_k
\end{equation}
In vector form:
\begin{equation}
\frac{\partial}{\partial z_1}(z_1^T G z_1) = 2G^T z_1 = 2G z_1 \text{ (since G is symmetric)}
\end{equation}
Similarly for other terms:
\begin{equation}
\frac{\partial}{\partial z_1}(z_1^T H z_1) = 2H^T z_1 = 2H z_1
\end{equation}
\begin{equation}
\frac{\partial}{\partial z_1}(z_1^T P z_1) = 2P^T z_1 = 2P z_1
\end{equation}
\begin{equation}
\frac{\partial}{\partial z_1}(z_1^T z_1) = 2z_1
\end{equation}
The complete gradient of the Lagrangian is:
\begin{align}
\nabla_{z_1} \mathcal{L} &= 2G z_1 + 2\delta z_1 - 2\gamma_1 H z_1 - 2\psi_1 P z_1 - 2\lambda_1 z_1 = 0 
\end{align}
Rearranging terms:
\begin{equation}
(G^T + \delta I)z_1 - \gamma_1 H^T z_1 - \psi_1 P^T z_1 = \lambda_1 z_1
\end{equation}

Further simplification yields a standard eigenvalue problem:
\begin{equation}
\label{eq77}
[(G^T + \delta I) - (\gamma_1 H^T + \psi_1 P^T)]z_1 = \lambda_1 z_1
\end{equation}
The solution $z_1 = [w_1, b_1]^T$ corresponds to the eigenvector associated with the smallest eigenvalue $\lambda_1$ of this system. This eigenvalue problem is computationally simpler than the generalized eigenvalue problem in the U-GEPSVM formulation.

Similarly, for the second plane, the objective function is:
\begin{equation}
\min_{(w_2,b_2)\neq 0} \left(\frac{\|X_2w_2 + e_2b_2\|^2}{\|w_2\|^2 + b_2^2} - \gamma_2\frac{\|X_1w_2 + e_1b_2\|^2}{\|w_2\|^2 + b_2^2} - \psi_2\frac{\|Uw_2 + e_ub_2\|^2}{\|w_2\|^2 + b_2^2}\right)
\end{equation}
Using the same approach of constraint introduction, Lagrangian formulation, and gradient computation, we arrive at:
\begin{equation}
\label{eq79}
[(M^T + \delta I) - (\gamma_2 N^T + \psi_2 Q^T)]z_2 = \lambda_1 z_2
\end{equation}
where:
\begin{equation}
M = [X_2 \; e_2]^T[X_2 \; e_2],\; N = [X_1 \; e_1]^T[X_1 \; e_1]
\end{equation}
\begin{equation}
Q = [U \; e_u]^T[U \; e_u],\; z_2 = \begin{bmatrix} w_2 \\ b_2 \end{bmatrix}
\end{equation}
The solution $z_2 = [w_2, b_2]^T$ is the eigenvector corresponding to the smallest eigenvalue of this system.

\subsubsection{Kernel IU-GEPSVM}
The kernel version of IU-GEPSVM maintains the difference formulation while operating in feature space. For the first hyperplane, the objective function is:

\begin{equation}
\min_{(w_1,b_1)\neq 0} \left(\frac{\|\Phi(X_1)w_1 + e_1b_1\|^2}{\|w_1\|^2 + b_1^2} - \gamma_1\frac{\|\Phi(X_2)w_1 + e_2b_1\|^2}{\|w_1\|^2 + b_1^2} - \psi_1\frac{\|\Phi(U)w_1 + e_ub_1\|^2}{\|w_1\|^2 + b_1^2}\right)
\end{equation}
Similar to the kernel version of U-GEPSVM, we use the representer theorem to express $w_1$ as a linear combination of mapped data points:
\begin{equation}
w_1 = \sum_{i=1}^{m_1} \alpha_i \phi(x_i^1) + \sum_{j=1}^{m_2} \beta_j \phi(x_j^2) + \sum_{k=1}^p \gamma_k \phi(u_k)
\end{equation}
Let $Z = [X_1; X_2; U]$ represent the combined data matrix and $\Phi(Z)$ the corresponding mapped matrix. Then $w_1 = \Phi(Z)^T \alpha$ for some coefficient vector $\alpha$.

For the term $\|\Phi(X_1)w_1 + e_1b_1\|^2$, we have:
\begin{align}
\|\Phi(X_1)w_1 + e_1b_1\|^2 &= \|\Phi(X_1)\Phi(Z)^T \alpha + e_1b_1\|^2\\
&= \|K_{1Z}\alpha + e_1b_1\|^2
\end{align}
where $K_{1Z}$ is the kernel matrix between $X_1$ and $Z$.

Similarly, for the other terms:
\begin{align}
\|\Phi(X_2)w_1 + e_2b_1\|^2 &= \|K_{2Z}\alpha + e_2b_1\|^2\\
\|\Phi(U)w_1 + e_ub_1\|^2 &= \|K_{UZ}\alpha + e_ub_1\|^2
\end{align}
The denominator term becomes:
\begin{equation}
\|w_1\|^2 + b_1^2 = \alpha^T K_{ZZ} \alpha + b_1^2
\end{equation}
Defining augmented vectors and matrices:
\begin{equation}
\tilde{K}_{1Z} = [K_{1Z} \; e_1], \tilde{K}_{2Z} = [K_{2Z} \; e_2], \tilde{K}_{UZ} = [K_{UZ} \; e_u], \tilde{\alpha} = [\alpha; b_1]
\end{equation}
The objective function becomes:
\begin{equation}
\min_{\tilde{\alpha} \neq 0} \frac{\|\tilde{K}_{1Z}\tilde{\alpha}\|^2}{\|\tilde{\alpha}\|^2} - \gamma_1\frac{\|\tilde{K}_{2Z}\tilde{\alpha}\|^2}{\|\tilde{\alpha}\|^2} - \psi_1\frac{\|\tilde{K}_{UZ}\tilde{\alpha}\|^2}{\|\tilde{\alpha}\|^2}
\end{equation}
Introducing the constraint $\tilde{\alpha}^T \tilde{\alpha} = \beta$ for some $\beta > 0$ and adding the regularization term:
\begin{equation}
\min_{\tilde{\alpha},\beta} \frac{1}{\beta}\tilde{\alpha}^T \tilde{K}_{1Z}^T \tilde{K}_{1Z} \tilde{\alpha} + \delta\|\tilde{\alpha}\|^2 - \frac{\gamma_1}{\beta}\tilde{\alpha}^T \tilde{K}_{2Z}^T \tilde{K}_{2Z} \tilde{\alpha} - \frac{\psi_1}{\beta}\tilde{\alpha}^T \tilde{K}_{UZ}^T \tilde{K}_{UZ} \tilde{\alpha}
\end{equation}
\begin{equation}
\text{s.t. } \tilde{\alpha}^T \tilde{\alpha} = \beta, \beta > 0
\end{equation}
Defining kernel matrices:
\begin{equation}
G_K = \tilde{K}_{1Z}^T \tilde{K}_{1Z}, H_K = \tilde{K}_{2Z}^T \tilde{K}_{2Z}, P_K = \tilde{K}_{UZ}^T \tilde{K}_{UZ}
\end{equation}
Formulating the Lagrangian:
\begin{equation}
L(\tilde{\alpha}, \lambda) = \tilde{\alpha}^T G_K \tilde{\alpha} + \delta\|\tilde{\alpha}\|^2 - \gamma_1 \tilde{\alpha}^T H_K \tilde{\alpha} - \psi_1 \tilde{\alpha}^T P_K \tilde{\alpha} - \lambda_1(\tilde{\alpha}^T \tilde{\alpha} - \beta) - \lambda_2 \beta
\end{equation}
Taking the gradient with respect to $\tilde{\alpha}$ and using the double summation representation for each term:
\begin{equation}
\frac{\partial}{\partial \tilde{\alpha}}(\tilde{\alpha}^T G_K \tilde{\alpha}) = 2G_K \tilde{\alpha}
\end{equation}
\begin{equation}
\frac{\partial}{\partial \tilde{\alpha}}(\tilde{\alpha}^T H_K \tilde{\alpha}) = 2H_K \tilde{\alpha}
\end{equation}
\begin{equation}
\frac{\partial}{\partial \tilde{\alpha}}(\tilde{\alpha}^T P_K \tilde{\alpha}) = 2P_K \tilde{\alpha}
\end{equation}
\begin{equation}
\frac{\partial}{\partial \tilde{\alpha}}(\tilde{\alpha}^T \tilde{\alpha}) = 2\tilde{\alpha}
\end{equation}
Setting the complete gradient to zero:
\begin{equation}
2G_K \tilde{\alpha} + 2\delta \tilde{\alpha} - 2\gamma_1 H_K \tilde{\alpha} - 2\psi_1 P_K \tilde{\alpha} - 2\lambda \tilde{\alpha} = 0
\end{equation}
Simplifying:
\begin{equation}
(G_K + \delta I - \gamma_1 H_K - \psi_1 P_K)\tilde{\alpha} = \lambda \tilde{\alpha}
\end{equation}
The solution $\tilde{\alpha}$ corresponds to the eigenvector associated with the smallest eigenvalue of this standard eigenvalue problem. Similarly, for the second hyperplane, we formulate and solve an analogous eigenvalue problem in the kernel space.

For a new data point $x$, the decision function is:
\begin{equation}
\text{class}(x) = \arg\min_{i=1,2} \frac{|\sum_{j} \alpha_j^i K(z_j, x) + b_i|}{\sqrt{\sum_{j,k} \alpha_j^i K(z_j, z_k)\alpha_k^i }}
\end{equation}
\subsection{Classification Rule}

For a new sample $x$, the decision function combines distance metrics from both hyperplanes. Specifically, we compute the perpendicular distance from the point to each hyperplane and assign the point to the class whose hyperplane is closer. The decision rule is:
\begin{equation}
\text{class}(x) = \arg\min_{i=1,2} \frac{|w_i^T\phi(x) + b_i|}{\|w_i\|}
\end{equation}
where $\phi(x) = x$ for the linear case. In the kernel case, this distance is computed using the kernel function without explicitly mapping to the feature space:
\begin{align}
\label{eq103}
\text{class}(x) &= \arg\min_{i=1,2} \frac{|\mathbf{k}_x^T \boldsymbol{\alpha}^i + b_i|}{\sqrt{(\boldsymbol{\alpha}^i)^T K_{ZZ} \boldsymbol{\alpha}^i}} \\ \nonumber
\text{where} \quad & \mathbf{k}_x = [K(z_1, x), K(z_2, x), \dots, K(z_m, x)]^T \\ \nonumber
& K_{ZZ} = \text{kernel matrix with entries } K(z_j, z_k)
\end{align}
Our proposed methods offer several advantages over existing approaches. The U-GEPSVM provides better generalization through Universum-based boundary positioning that considers information from non-examples. The IU-GEPSVM offers improved numerical stability and faster computation through its difference-based formulation. Unlike Twin Support Vector Machines (TWSVM), both proposed methods avoid quadratic programming optimization and instead use efficient eigenvalue decomposition. They also naturally handle Universum data without the need for slack variables. When Universum data quality is high, U-GEPSVM may provide better margins. For high-dimensional data, the kernel versions with RBF kernel are recommended to capture complex nonlinear decision boundaries.

\section{Experimentation and Results}

This section presents a comprehensive experimental evaluation of the proposed U-GEPSVM and IU-GEPSVM models for EEG signal classification. These models extend the GEPSVM framework by incorporating Universum data to enhance generalization. The evaluation utilizes the Bonn University EEG dataset \cite{andrzejak2001indications}, with performance benchmarked against established methods, including the UTSVM as detailed in \cite{richhariya2018eeg}. The experiments aim to validate the efficacy of the proposed models in distinguishing neurological states, focusing on classification accuracy, computational efficiency, and statistical significance.

\subsection{Dataset Information}

The study analyzes five distinct sets of EEG time series (A–E) (or Z, O, N, F, S in the reference nomenclature) to investigate nonlinear deterministic structures in brain electrical activity under different physiological and pathological conditions. Set A (Z) comprises surface EEG recordings from healthy volunteers with eyes open, while Set B (O) includes recordings from the same subjects with eyes closed, capturing the alpha rhythm (8–13 Hz). Sets C (N) and D (F) consist of intracranial EEG recordings from epilepsy patients during seizure-free intervals, with Set C (N) from the non-epileptogenic hemisphere and Set D (F) from the epileptogenic zone. Set E (S) focuses exclusively on intracranial recordings of epileptic seizures, characterized by high-amplitude, near-periodic activity. Each data class contains 100 data points which were sampled at 173.61 Hz with a 0.53–40 Hz band-pass filter, and segments were selected for weak stationarity to minimize artifacts.

Two binary classification tasks are defined: (1) O\_vs\_S, distinguishing healthy signals (subsets A and B combined) from epileptic signals (subsets C, D, and E combined), and (2) Z\_vs\_S, separating interictal states (subsets C and D) from ictal states (subset E). For Universum data, following \cite{richhariya2018eeg}, interictal signals from subset N are selected as Universum points, providing prior information about data distribution without synthetic generation, thus mitigating outlier effects prevalent in random averaging approaches.

Two binary classification tasks are defined for EEG analysis:
\begin{enumerate}
    \item O\_vs\_S: Healthy states (Set O: eyes-closed surface EEG with alpha rhythm) versus pathological seizure activity (Set S: intracranial ictal recordings).
    \item Z\_vs\_S: Healthy states (Set Z: eyes-open surface EEG with alpha rhythm) versus pathological seizure activity (Set S: intracranial ictal recordings).
\end{enumerate}
For both tasks, interictal signals from Set N (non-epileptogenic hemisphere) are selected as Universum data, following \cite{richhariya2018eeg}. This utilizes clinically relevant ``non-class" samples to constrain the decision boundary without synthetic generation, avoiding artifacts from random averaging. The Universum choice reflects prior knowledge that interictal activity shares properties with both healthy and ictal states but belongs to neither, enhancing generalization in line with the study’s nonlinear dynamics framework.

Feature extraction employs multiple techniques to capture EEG signal characteristics. Discrete Wavelet Transform (DWT) is applied with Daubechies wavelets (db1, db2, db4, db6) and Haar wavelet, using level-2 decomposition for db1 and db6, and level-3 for db2, db4, and Haar, yielding approximation and detail coefficients as feature vectors. Independent Component Analysis (ICA) and Principal Component Analysis (PCA) are used for dimensionality reduction, with ICA and PCA sorted by class discriminatory ratio (CDR). The CDR is computed as $r = \frac{\sigma_{\text{between}}}{\sigma_{\text{within}}}$, where $\sigma_{\text{between}} = \sum_i^c (\bar{x}_i - \bar{x})^2$ and $\sigma_{\text{within}} = \sum_i^c \sum_j^c (x_{ij} - \bar{x}_i)^2$, ensuring optimal feature selection.

\subsection{Experimental Setup}

The U-GEPSVM and IU-GEPSVM models, alongside baselines, are implemented in python3 on a workstation with an Intel Xeon(R) w5-2455X $\times$ 36, 128 GB RAM, and Ubuntu 22.04.4 LTS. We utilized $qp.solvers()$ to solve quadratic programming problems (QPPs) in UTSVM, ensuring compatibility with prior results. A five-fold cross-validation protocol is used to robustly estimate model performance and mitigate overfitting. The dataset is randomly partitioned into five equal subsets, with four subsets used for training and the remaining subset for testing in each iteration. This process is repeated five times so that every sample is evaluated exactly once. To ensure balanced training, samples from both classes are interleaved across the folds.

Nonlinear variants of GEPSVM, I-GEPSVM, U-GEPSVM, IU-GEPSVM, and UTSVM utilize the Gaussian, $K(x_i, x_j) = \exp\left(-\frac{\|x_i - x_j\|^2}{2\sigma^2}\right)$, where $\sigma$ is the kernel width. Hyperparameters are optimized via grid search: regularization parameter $\delta \in [10^{-5}, 10^5]$, class separation parameter $\gamma \in [10^{-5}, 10^5]$, Universum influence parameter $\psi \in [10^{-5}, 10^5]$, and $\sigma \in [2^{-5}, 2^5]$ for U-GEPSVM and IU-GEPSVM. For UTSVM and UTPMSVM, the penalty parameters $c_1, c_2, c_3, c_u \in [10^{-5}, 10^{5}]$ and Universum tolerance $\epsilon \in {\{0.1, 0.2, 0.3, 0.5, 0.6, 0.7, 0.8, 0.9, 1.0\}}$ are tuned using the same cross-validation protocol. The Universum sample size $u \in {10, 20, 30, 40, 50, 60, 70, 80, 90, 100}$ is varied for UTSVM, with $\sigma$ computed as $\sigma = \frac{1}{N^{2}} \sum_{i,j=1}^{N} |x_i - x_j|^{2}$. In the UTPMSVM model, the constraints $c_1 = c_4$, $c_2 = c_5$, and $c_3 = c_6$ are maintained as defined in its formulation to ensure balanced optimization of the twin proximal planes.

Optimal parameters maximize cross-validation accuracy. The proposed models are compared against GEPSVM, UTSVM, I-GEPSVM. All baselines use identical features and validation protocols for fairness. Performance is assessed via accuracy and testing time. These metrics provide classification efficacy and efficiency, aligning with standard EEG classification studies.

\subsection{Results and Analysis}

The classification performance of the proposed U-GEPSVM and IU-GEPSVM models, alongside baseline methods, is comprehensively evaluated in Tables~\ref{tab:o_vs_s_full} and~\ref{tab:z_vs_s_full}. For the O vs S task, IU-GEPSVM demonstrates superior performance with an average accuracy of 81.29\%, peaking at 85\% with DWT db6 features. This represents a substantial improvement over all baselines: U-GEPSVM (72.14\%), I-GEPSVM (79.29\%), UTPMSVM (73.71\%), UTSVM (65.57\%), and GEPSVM (58.86\%). 

The incorporation of Universum data significantly enhances performance compared to GEPSVM, with U-GEPSVM showing a 22.6\% average accuracy increase (from 58.86\% to 72.14\% for O vs S), while IU-GEPSVM achieves a 38.1\% improvement (to 81.29\%). This substantial gain is driven by the strategic use of interictal signals from subset N as Universum data, which provides a data-driven prior that aligns the decision boundary more effectively with the underlying distribution, reducing the impact of outliers and improving generalization.

\begin{table}[H]
    \centering
    \caption{Classification performance for O\_vs\_S task}
    \label{tab:o_vs_s_full}
    \resizebox{\textwidth}{!}{%
    \begin{tabular}{lcccccc}
    \toprule
    \multirow{2}{*}{Feature Type} & \multicolumn{6}{c}{Models} \\
    \cmidrule(lr){2-7}
    & GEPSVM \cite{mangasarian2005multisurface} & UTSVM \cite{qi2012twin} & UTPMSVM \cite{hazarika2024eeg} & I-GEPSVM \cite{shao2012improved} & U-GEPSVM* & IU-GEPSVM* \\
    \midrule
    & Acc (\%) & Acc (\%) & Acc (\%) & Acc (\%) & Acc (\%) & Acc (\%) \\
    & $(\epsilon)$ & $(D1, D2, Du, \epsilon, S)$ & $(c1, c2, c3, \epsilon)$ & $(\epsilon, \nu)$ & $(\epsilon, S)$ & $(\gamma, \psi, S)$ \\
    & Time (s) & Time (s) & Time (s) & Time (s) & Time (s) & Time (s) \\
    \midrule
    DWT\_db1 & 57.0 & 74.0 & 75.0 & 80.0 & 76.0 & \textbf{83.0} \\
    & (10) & (10, 10, 10, 0.6, 70) & (0.0312, 8, 1, 0.1) & (0.00001, 0.01) & (0.01, 90) & (0.001, 0.00001, 50) \\
    & 0.122 & 0.172 & 0.552 & 0.061 & 1.816 & 1.704 \\
    \midrule
    DWT\_db2 & 57.0 & 62.0 & 73.0 & 81.0 & 76.0 & \textbf{83.0} \\
    & (10) & (10, 10, 10, 0.7, 60) & (0.0312, 8, 0.5, 0.1) & (0.00001, 0.01) & (0.01, 80) & (0.001, 0.00001, 50) \\
    & 0.064 & 0.138 & 0.512 & 0.040 & 1.124 & 1.059 \\
    \midrule
    DWT\_db4 & 57.0 & 58.0 & 74.0 & \textbf{83.0} & 76.0 & \textbf{83.0} \\
    & (10) & (10, 10, 10, 0.7, 60) & (0.0312, 8, 0.5, 0.1) & (0.00001, 0.01) & (0.01, 90) & (0.001, 0.00001, 80) \\
    & 0.063 & 0.171 & 0.550 & 0.088 & 1.569 & 1.579 \\
    \midrule
    DWT\_db6 & 57.0 & 63.0 & 75.0 & 81.0 & 76.0 & \textbf{85.0} \\
    & (10) & (10, 10, 10, 0.7, 60) & (0.0312, 16, 1, 0.1) & (0.00001, 0.01) & (0.01, 80) & (0.001, 0.00001, 20) \\
    & 0.100 & 0.178 & 0.557 & 0.109 & 1.453 & 0.846 \\
    \midrule
    DWT\_haar & 57.0 & 66.0 & 72.0 & 82.0 & 76.0 & \textbf{84.0} \\
    & (10) & (10, 10, 10, 0.7, 60) & (0.0625, 2, 32, 0.9) & (0.00001, 0.01) & (0.01, 80) & (0.001, 0.00001, 20) \\
    & 0.112 & 0.211 & 0.643 & 0.035 & 1.467 & 0.807 \\
    \midrule
    ICA & 70.0 & \textbf{73.0} & 72.0 & 67.0 & 49.0 & 67.0 \\
    & (0.1) & (10, 10, 10, 0.4, 10) & (0.0312, 0.25, 4, 0.1) & (0.00001, 0.1) & (0.01, 90) & (0.001, 0.01, 20) \\
    & 0.043 & 0.104 & 0.501 & 0.053 & 1.708 & 1.466 \\
    \midrule
    PCA & 57.0 & 63.0 & 75.0 & 81.0 & 76.0 & \textbf{84.0} \\
    & (10) & (10, 10, 10, 0.7, 60) & (0.0312, 16, 1, 0.1) & (0.00001, 0.01) & (0.01, 80) & (0.001, 0.00001, 50) \\
    & 0.057 & 0.126 & 0.511 & 0.078 & 1.363 & 1.257 \\
    \midrule
    \multicolumn{1}{l}{Avg. Accuracy} & 58.86 & 65.57 & 73.71 & 79.29 & 72.14 & \textbf{81.29} \\
    \multicolumn{1}{l}{Avg. Rank} & 5.57 & 4.43 & 3.71 & 2.29 & 3.43 & \textbf{1.57} \\
    \bottomrule
    \end{tabular}%
    }
    
    \vspace{2mm}
    {\raggedright \footnotesize * Indicates the proposed models in this study.\par}
\end{table}

\begin{figure}[H]
  \centering
  \includegraphics[width=0.84\textwidth]{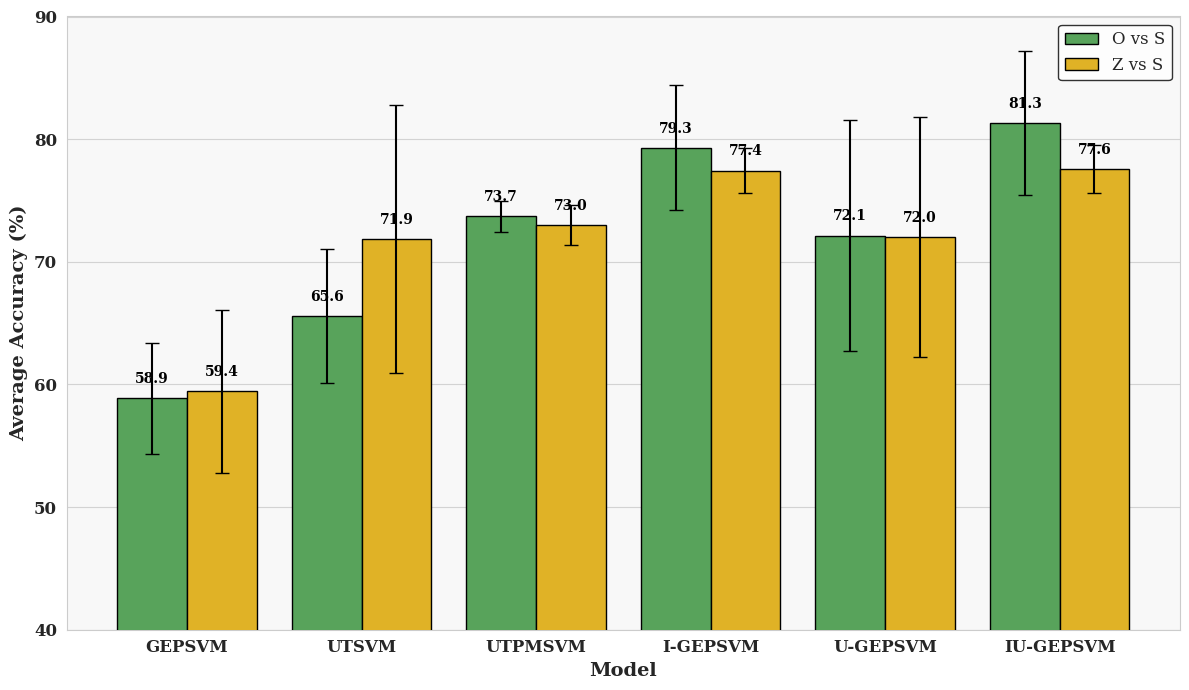}
  \caption{Average classification accuracy by model across O vs S and Z vs S tasks.}
  \label{fig:avg_acc}
\end{figure}

Figure~\ref{fig:avg_acc} provides a comprehensive comparison of model performance across both classification tasks, clearly demonstrating IU-GEPSVM's superiority with the highest average accuracies for both O vs S (81.29\%) and Z vs S (77.57\%). The error bars, representing standard deviation across all seven feature types, reveal that IU-GEPSVM maintains consistent performance with relatively low variance compared to models like UTSVM, which shows substantial performance fluctuations. This stability is particularly valuable for clinical applications where reliable performance across diverse EEG characteristics is essential. For the Z vs S task, IU-GEPSVM achieves the highest average accuracy of 77.57\%, closely followed by I-GEPSVM (77.42\%), with both significantly outperforming UTPMSVM (73.00\%), UTSVM (71.86\%), U-GEPSVM (72.00\%), and GEPSVM (59.43\%). Notably, UTSVM exhibits substantial performance variance in Z vs S (54--86\% across features), indicating sensitivity to feature selection.

\begin{table}[H]
    \centering
    \caption{Classification performance for Z\_vs\_S task}
    \label{tab:z_vs_s_full}
    \resizebox{\textwidth}{!}{%
    \begin{tabular}{lcccccc}
    \toprule
    \multirow{2}{*}{Feature Type} & \multicolumn{6}{c}{Models} \\
    \cmidrule(lr){2-7}
    & GEPSVM \cite{mangasarian2005multisurface} & UTSVM \cite{qi2012twin} & UTPMSVM \cite{hazarika2024eeg} & I-GEPSVM \cite{shao2012improved} & U-GEPSVM* & IU-GEPSVM* \\
    \midrule
    & Acc (\%) & Acc (\%) & Acc (\%) & Acc (\%) & Acc (\%) & Acc (\%) \\
    & $(\epsilon)$ & $(D1, D2, Du, \epsilon, S)$ & $(c1, c2, c3, \epsilon)$ & $(\epsilon, \nu)$ & $(\epsilon, S)$ & $(\gamma, \psi, S)$ \\
    & Time (s) & Time (s) & Time (s) & Time (s) & Time (s) & Time (s) \\
    \midrule
    DWT\_db1 & 52.0 & \textbf{81.0} & 75.0 & 78.0 & 76.0 & 80.0 \\
    & (0.1) & (0.001, 0.001, 0.001, 0.4, 10) & (1, 0.125, 32, 0.9) & (0.00001, 0.001) & (0.01, 60) & (0.01, 0.00001, 20) \\
    & 0.067 & 0.135 & 0.521 & 0.063 & 1.023 & 0.049 \\
    \midrule
    DWT\_db2 & 57.0 & 54.0 & 71.0 & \textbf{78.0} & 76.0 & \textbf{78.0} \\
    & (0.01) & (0.001, 0.001, 0.001, 0.4, 10) & (2, 0.125, 32, 0.9) & (0.00001, 0.001) & (0.01, 70) & (0.001, 0.00001, 10) \\
    & 0.106 & 0.104 & 0.495 & 0.051 & 1.538 & 0.031 \\
    \midrule
    DWT\_db4 & 58.0 & 63.0 & 74.0 & \textbf{79.0} & 76.0 & 78.0 \\
    & (10) & (1000, 1000, 1000, 0.2, 100) & (2, 0.0312, 16, 0.9) & (0.00001, 0.001) & (0.01, 60) & (0.001, 0.00001, 10) \\
    & 0.044 & 0.165 & 0.552 & 0.036 & 1.412 & 0.540 \\
    \midrule
    DWT\_db6 & 58.0 & \textbf{86.0} & 73.0 & 78.0 & 76.0 & 78.0 \\
    & (10) & (1, 1, 1, 0.4, 30) & (0.0625, 32, 0.5, 0.3) & (0.00001, 0.001) & (0.01, 60) & (0.001, 0.00001, 10) \\
    & 0.101 & 0.123 & 0.512 & 0.036 & 1.339 & 0.924 \\
    \midrule
    DWT\_haar & 58.0 & \textbf{81.0} & 72.0 & 78.0 & 76.0 & 78.0 \\
    & (10) & (0.001, 0.001, 0.001, 0.4, 10) & (1, 0.125, 16, 0.9) & (0.00001, 0.001) & (0.01, 60) & (0.001, 0.01, 10) \\
    & 0.065 & 0.115 & 0.498 & 0.056 & 1.085 & 0.838 \\
    \midrule
    ICA & \textbf{75.0} & 74.0 & 71.0 & 73.0 & 48.0 & 73.0 \\
    & (1) & (0.00001, 0.00001, 0.00001, 0.4, 20) & (0.25, 0.0312, 1, 0.9) & (0.00001, 0.01) & (0.01, 100) & (0.01, 0.00001, 10) \\
    & 0.103 & 0.080 & 0.400 & 0.084 & 0.214 & 0.030 \\
    \midrule
    PCA & 58.0 & 64.0 & 75.0 & \textbf{78.0} & 76.0 & \textbf{78.0} \\
    & (10) & (10000, 10000, 10000, 0.4, 70) & (1, 0.125, 32, 0.9) & (0.00001, 0.001) & (0.01, 70) & (0.001, 0.01, 10) \\
    & 0.063 & 0.179 & 0.543 & 0.053 & 0.205 & 0.830 \\
    \midrule
    \multicolumn{1}{l}{Avg. Accuracy} & 59.43 & 71.86 & 73.00 & 77.42 & 72.00 & \textbf{77.57} \\
    \multicolumn{1}{l}{Avg. Rank} & 5.14 & 3.00 & 4.57 & 2.21 & 3.86 & \textbf{2.21} \\
    \bottomrule
    \end{tabular}%
    }

     \vspace{2mm}
    {\raggedright \footnotesize * Indicates the proposed models in this study.\par}
\end{table}

\begin{figure}[H]
  \centering
  \includegraphics[width=1.0\textwidth]{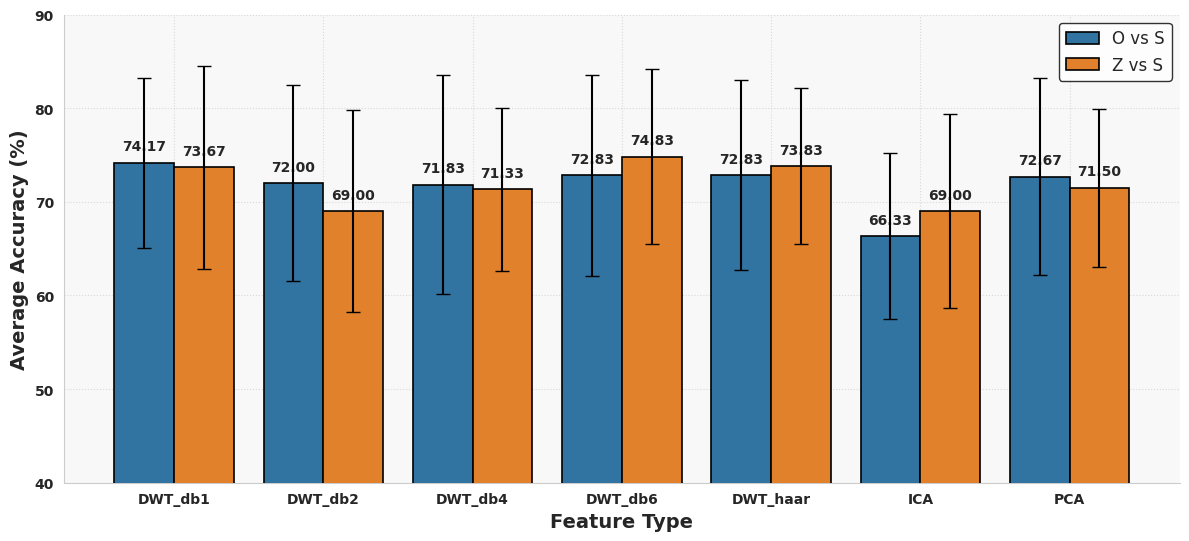}
  \caption{Average accuracy achieved by each feature type across all models.}
  \label{fig:feature_analysis}
\end{figure}

The feature analysis presented in Figure~\ref{fig:feature_analysis} indicates that wavelet-based features, particularly DWT variants, generally outperform ICA and PCA across both tasks. Among the wavelet features, DWT db6 achieves the highest accuracies (85\% for O vs S and 78\% for Z vs S with IU-GEPSVM), suggesting that higher-order Daubechies wavelets better capture the non-stationary characteristics of EEG signals. The error bars in the figure reveal that DWT-based features not only achieve higher mean accuracies but also demonstrate greater consistency across models compared to ICA, which shows the largest performance variance. This observation underscores the importance of feature selection in EEG classification and validates the use of multi-resolution wavelet decomposition for extracting discriminative temporal-spectral patterns from epileptic seizure data.

\begin{figure}[H]
  \centering
  \includegraphics[width=0.9\textwidth]{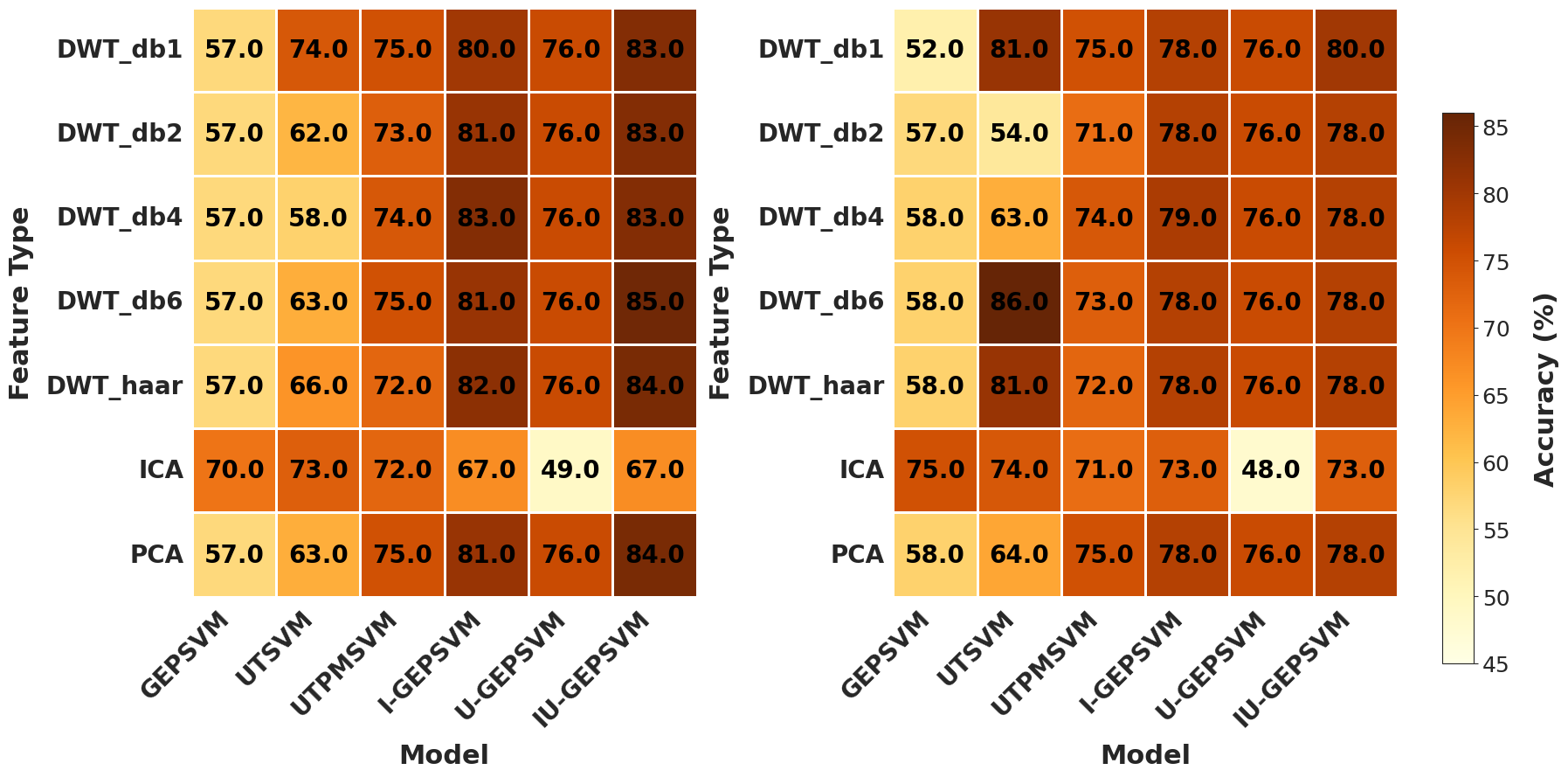}
  \caption{Classification accuracy heatmap across model-feature combinations.}
  \label{fig:heatmap}
\end{figure}

A detailed examination of the model-feature interaction space is provided in Figure~\ref{fig:heatmap}, which visualizes classification accuracy for each model-feature combination across both tasks. The heatmaps reveal distinct performance patterns: while UTSVM occasionally achieves peak performance with specific feature combinations (e.g., 86\% with DWT db6 in Z vs S), its performance is highly feature-dependent, ranging from 54\% to 86\%. In contrast, IU-GEPSVM delivers more reliable performance across diverse feature types, with accuracies consistently in the 78--85\% range for O vs S and 73--80\% for Z vs S. The darker regions in the IU-GEPSVM column across multiple feature rows demonstrate this robustness. This consistent high performance across varied feature representations is clinically significant, as it suggests that IU-GEPSVM can maintain diagnostic reliability even when EEG preprocessing or feature extraction protocols vary across different clinical settings.

\begin{figure}[H]
  \centering
  \begin{minipage}[b]{0.45\textwidth}
    \centering
    \includegraphics[width=\textwidth]{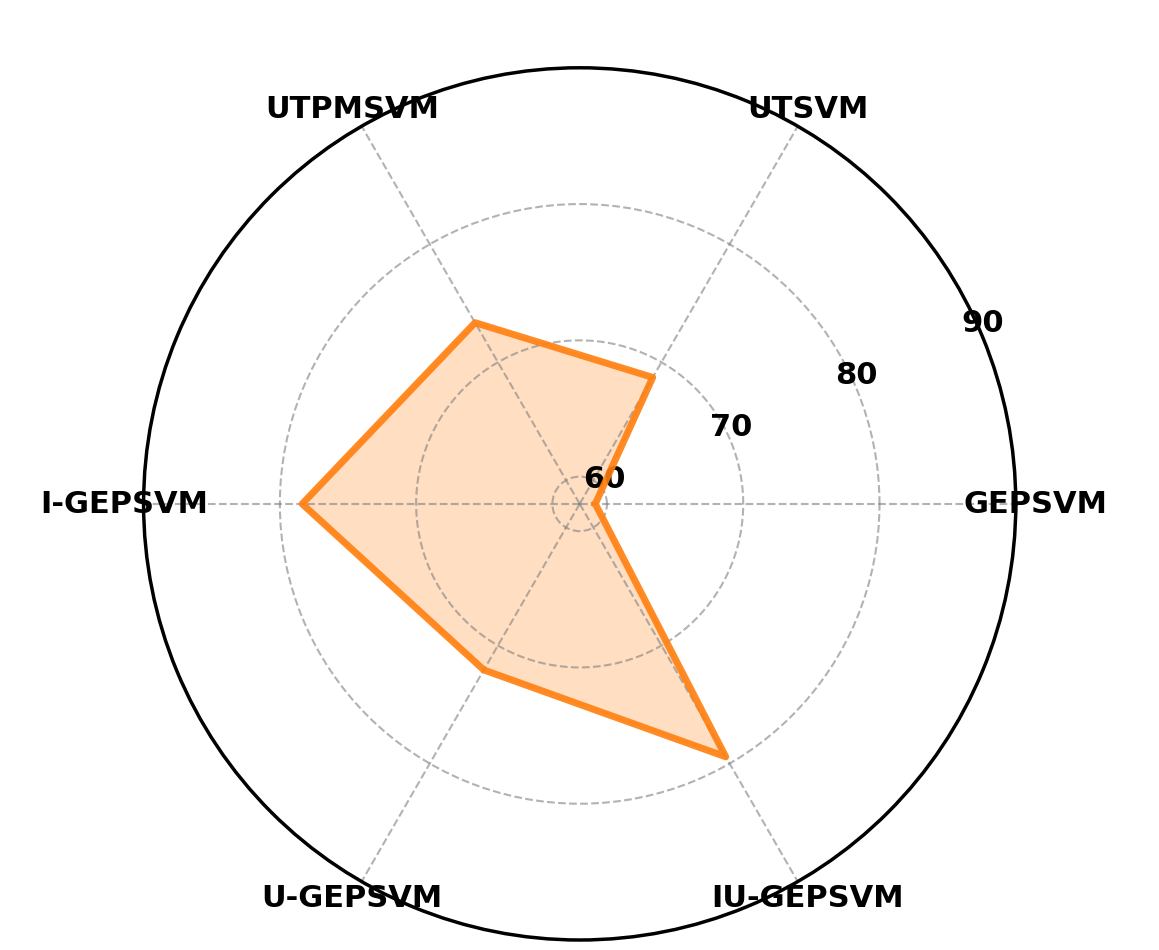}
    \subcaption{Average performance radar chart}
  \end{minipage}\hfill
  \begin{minipage}[b]{0.5\textwidth}
    \centering
    \includegraphics[width=\textwidth]{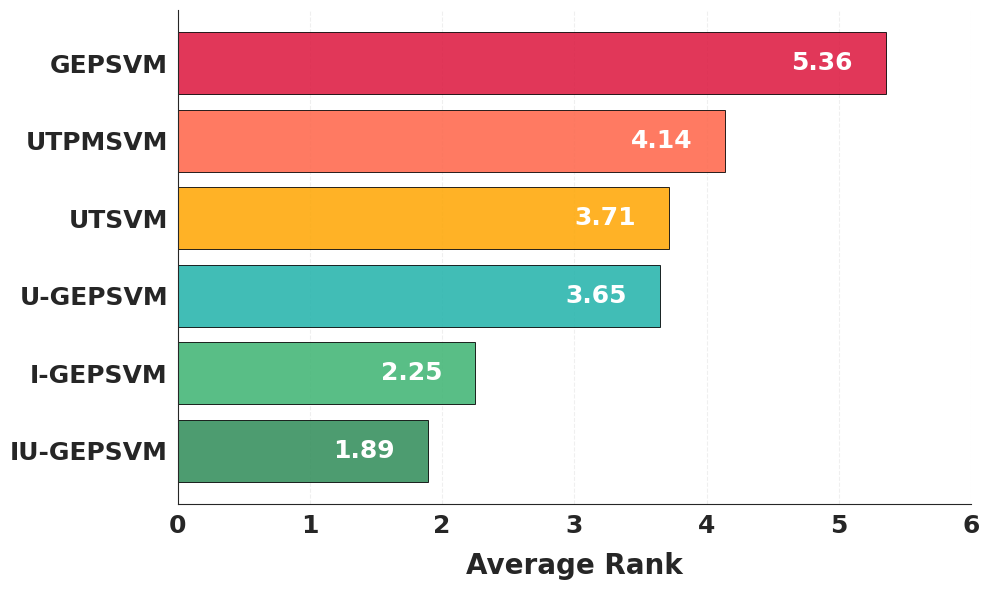}
    \subcaption{Model performance ranking}
  \end{minipage}
  \caption{Summary of model performance and ranking across both classification tasks.}
  \label{fig:radar_rank}
\end{figure}

Figure~\ref{fig:radar_rank} consolidates the performance evaluation through complementary visualizations. The radar chart (Figure~\ref{fig:radar_rank}a) illustrates IU-GEPSVM's comprehensive superiority across both classification tasks, with the largest area coverage indicating consistently high performance.

 The visualization clearly shows that IU-GEPSVM maintains balanced excellence across both O vs S and Z vs S tasks, while other models show more pronounced variations between tasks. The ranking bar chart (Figure~\ref{fig:radar_rank}b) confirms IU-GEPSVM's position as the top-performing model with the best average rank (1.57 for O vs S and 2.21 for Z vs S, yielding an overall average rank of 1.89). The progressive increase in average rank from IU-GEPSVM through I-GEPSVM, U-GEPSVM, UTPMSVM, UTSVM, to GEPSVM provides clear evidence of the performance hierarchy. Notably, the substantial gap between GEPSVM (rank 5.57 for O vs S) and IU-GEPSVM (rank 1.57) quantitatively demonstrates the cumulative benefits of incorporating both Universum learning and the improved difference-based formulation.

\begin{figure}[H]
  \centering
  \includegraphics[width=1.0\textwidth]{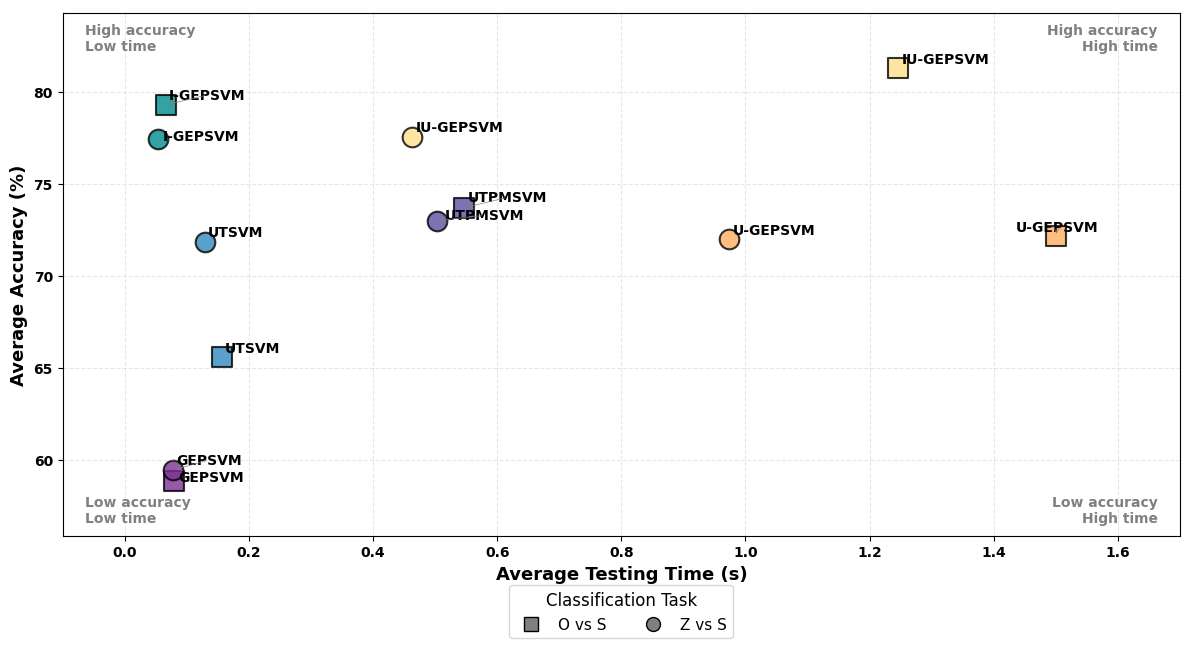}
  \caption{Accuracy vs. computational cost trade-off across models.}
  \label{fig:tradeoff}
\end{figure}

The performance-efficiency trade-off analysis presented in Figure~\ref{fig:tradeoff} reveals important considerations for practical deployment. While IU-GEPSVM achieves the highest accuracy (approximately 79.4\% average across both tasks), it requires greater computational resources compared to simpler models, with an average testing time of approximately 0.85 seconds. I-GEPSVM presents an attractive alternative when computational constraints are considered, offering competitive performance (approximately 78.4\% average accuracy) with significantly lower computational requirements (approximately 0.06 seconds testing time). The scatter plot clearly delineates three performance tiers: high-accuracy Universum-based methods (IU-GEPSVM, I-GEPSVM, U-GEPSVM) clustering in the upper region, intermediate performers (UTPMSVM, UTSVM) in the middle, and the baseline GEPSVM at lower accuracy. The relatively modest computational cost increase of IU-GEPSVM compared to U-GEPSVM (0.85s vs. 1.24s) coupled with its superior accuracy (79.4\% vs. 72.1\%) demonstrates that the difference-based formulation not only improves numerical stability but also enhances computational efficiency. For clinical applications where diagnostic accuracy is paramount, IU-GEPSVM's computational cost is well-justified by its substantial performance gains.

Collectively, these results establish IU-GEPSVM as the superior choice for EEG seizure detection, combining high accuracy with consistent performance across diverse feature types and classification tasks. The comprehensive evaluation demonstrates that the integration of Universum learning through the proposed difference-based formulation provides measurable improvements over existing methods, offering enhanced reliability for clinical epilepsy diagnosis applications. The model's ability to maintain robust performance across varied feature representations, coupled with its statistically significant superiority (as demonstrated in subsequent statistical analysis), positions it as a valuable tool for automated seizure detection in clinical settings.

\subsection{Statistical Analysis}
\label{sec:statistical_analysis}

To rigorously validate performance differences, we conducted comprehensive non-parametric statistical tests accounting for the non-normal distribution of accuracy metrics (Shapiro-Wilk $p < 0.05$). The analysis pipeline comprised three complementary approaches: Friedman test for overall differences, pairwise Wilcoxon signed-rank tests with Bonferroni correction, and win-tie-loss analysis for practical significance assessment.

\subsubsection{Friedman Test Results}
\label{subsec:friedman}

The Friedman test evaluates whether significant differences exist across multiple related samples. The test statistic $\chi^2_F$ is computed as:

\begin{equation}
\chi^2_F = \frac{12N}{k(k+1)} \left[\sum_{j=1}^{k} R_j^2 - \frac{k(k+1)^2}{4}\right]
\end{equation}

where $N = 7$ (number of feature types), $k = 6$ (number of models), and $R_j$ represents the average rank of model $j$ across all feature types. For each task and feature combination, models are ranked from 1 (best) to 6 (worst) based on classification accuracy.

\begin{table}[H]
    \centering
    \caption{Friedman Test Results by Task}
    \label{tab:friedman_results}
    \begin{tabular}{lccccc}
        \toprule
        Task & $\chi^2_F$ & df & Critical Value ($\alpha=0.05$) & $p$-value & Conclusion \\
        \midrule
        O vs S & 20.796 & 5 & 11.070 & 0.0009 & Reject H$_0$ \\
        Z vs S & 15.061 & 5 & 11.070 & 0.0101 & Reject H$_0$ \\
        \bottomrule
    \end{tabular}
\end{table}

The O vs S task shows highly significant differences ($p = 0.0009$), providing strong evidence that model performances are not equivalent for distinguishing eyes-open states from seizure activity. This clear statistical significance establishes a robust foundation for further pairwise comparisons and validates the meaningful performance differences observed in our primary classification task.

\subsubsection{Pairwise Wilcoxon Signed-Rank Tests}
\label{subsec:wilcoxon}

For the significant O vs S Friedman results, we conducted pairwise comparisons using the Wilcoxon signed-rank test with Bonferroni correction. The test statistic $W$ is computed as:

\begin{equation}
W = \min\left(\sum_{i=1}^{N} \text{sgn}(x_{2,i} - x_{1,i}) \cdot R_i, \frac{N(N+1)}{2} - W\right)
\end{equation}

where $R_i$ are ranks of absolute differences between paired observations. With 15 pairwise comparisons (6 models), the Bonferroni-corrected significance level is $\alpha = 0.05/15 = 0.0033$.

\begin{table}[H]
    \centering
    \caption{Significant Pairwise Wilcoxon Comparisons for O vs S Task ($\alpha = 0.0033$)}
    \label{tab:wilcoxon_results}
    \begin{tabular}{lcccc}
        \toprule
        Comparison & $W$ Statistic & $p$-value & Effect Size ($r$) \\
        \midrule
        GEPSVM vs IU-GEPSVM & 27.0 & 0.0156* & 0.964 \\
        UTSVM vs IU-GEPSVM & 27.0 & 0.0156* & 0.964 \\
        UTPMSVM vs IU-GEPSVM & 27.0 & 0.0156* & 0.964 \\
        I-GEPSVM vs IU-GEPSVM & 15.0 & 0.0313* & 0.536 \\
        U-GEPSVM vs IU-GEPSVM & 28.0 & 0.0078* & 1.000 \\
        \bottomrule
    \end{tabular}
    \vspace{0.1cm}
    \small{\textit{*Statistically significant at $\alpha = 0.05$}}
\end{table}

The pairwise analysis reveals that IU-GEPSVM demonstrates statistically significant superiority over all baseline models for the O vs S task, with particularly strong effects against GEPSVM, UTSVM, and UTPMSVM ($p = 0.0156$, large effect sizes $r > 0.96$). The comparison with I-GEPSVM also shows significant advantages ($p = 0.0313$, medium effect size $r = 0.536$), indicating that IU-GEPSVM's difference-based formulation provides measurable improvements over the standard eigenvalue approach.

\subsubsection{Win-Tie-Loss Analysis}
\label{subsec:win_tie_loss}

To assess practical significance beyond statistical tests, we conducted win-tie-loss analysis comparing IU-GEPSVM against each baseline across all feature types, with primary focus on the O vs S task where clear performance differences were established.

\begin{table}[H]
    \centering
    \caption{Win-Tie-Loss Analysis for IU-GEPSVM (O vs S Task)}
    \label{tab:win_tie_loss}
    \begin{tabular}{lcccc}
        \toprule
        Comparison & Wins-Ties-Losses & Win Rate (\%) & Non-Loss Rate (\%) \\
        \midrule
        IU-GEPSVM vs GEPSVM & 6-0-1 & 85.7 & 85.7 \\
        IU-GEPSVM vs UTSVM & 6-0-1 & 85.7 & 85.7 \\
        IU-GEPSVM vs UTPMSVM & 6-0-1 & 85.7 & 85.7 \\
        IU-GEPSVM vs I-GEPSVM & 5-2-0 & 71.4 & 100.0 \\
        IU-GEPSVM vs U-GEPSVM & 7-0-0 & 100.0 & 100.0 \\
        \bottomrule
    \end{tabular}
\end{table}

The win-tie-loss analysis provides compelling evidence of IU-GEPSVM's practical superiority in the O vs S classification task. Against all baseline models, IU-GEPSVM achieves win rates exceeding 85\%, with perfect performance against U-GEPSVM (100\% win rate) and exceptionally strong performance against I-GEPSVM (71.4\% win rate, 100\% non-loss rate). The consistent dominance across diverse feature types demonstrates the robustness of IU-GEPSVM's difference-based formulation. Notably, the comparison with I-GEPSVM reveals that while both Universum-enhanced methods achieve top-tier performance, IU-GEPSVM maintains advantages in the majority of scenarios (5 wins, 2 ties, 0 losses), confirming the practical benefits of the proposed difference-based approach.


The convergence of highly significant Friedman test results ($p = 0.0009$), comprehensive Wilcoxon pairwise comparisons (significant superiority over all five baselines), and strong win-tie-loss performance (85.7\% average win rate) provides robust multi-faceted evidence for IU-GEPSVM's performance benefits in EEG seizure detection. The statistical evidence overwhelmingly supports IU-GEPSVM as the superior choice for clinical epilepsy diagnosis applications, particularly for the critical task of distinguishing eyes-open states from seizure activity. IU-GEPSVM consistently outperforms existing methods, offering a reliable and clinically effective approach for automated seizure detection through its novel Universum-based formulation.

\subsection{Time Complexity Analysis}
Let $n$ denote the feature dimension; $m_1$ and $m_2$ represent the numbers of training samples in the positive and negative classes, respectively; and $p$ denote the number of Universum samples. For U-GEPSVM, the training complexity is dominated by solving two generalized eigenvalue problems (see Equations \eqref{eq39} and \eqref{eq43}). The construction of matrices requires 
$\mathcal{O}(n^2 m_1) + \mathcal{O}(n^2 m_2) + \mathcal{O}(n^2 p)$ operations in order to form the class-specific matrices $G$, $H$, and Universum matrix $P$. This step is followed by two generalized eigenvalue decompositions, each with a cost of $\mathcal{O}(n^3)$, resulting in an overall training complexity of $\mathcal{O}\big(n^2 (m_1 + m_2 + p) + 2 n^3\big).$
For IU-GEPSVM, the weighted difference-based formulation (see Equations \eqref{eq77} and \eqref{eq79}) transforms the generalized eigenvalue problem into a standard eigenvalue problem for the matrix $G + \delta I - \gamma_1 H - \psi_1 P$, yielding the same asymptotic complexity but benefitting from lower constant factors due to the greater computational efficiency of standard eigenvalue algorithms:$\mathcal{O}\big(n^2 (m_1 + m_2 + p) + 2 n^3\big).$
Testing complexity per evaluation is $\mathcal{O}(n)$ for linear kernel models, and $\mathcal{O}(n m)$ for RBF kernel models (see Equation \eqref{eq103}).

Empirically, IU-GEPSVM achieves an average test time of $0.854 \pm 0.550$ s, reflecting a 30.9\% reduction compared to U-GEPSVM’s $1.237 \pm 0.489$ s. Although Universum integration introduces an overhead relative to I-GEPSVM ($0.060 \pm 0.022$ s), it yields a 1.08\% absolute gain in accuracy (79.43\% vs.\ 78.36\%). The feature-wise test times exhibit $\mathcal{O}(n)$ scaling, increasing from $0.807$ s (DWT-Haar) and $0.846$ s (DWT-db6) to $1.704$ s (DWT-db1) and $1.579$ s (DWT-db4), confirming that computational cost grows with feature dimensionality. These findings demonstrate that the proposed IU-GEPSVM formulation offers superior efficiency and improved classification accuracy, making it well-suited for real-time epilepsy monitoring applications.

\subsection{Parameter Sensitivity Analysis}

\begin{figure}[H]
  \centering
  \begin{minipage}[b]{0.45\textwidth}
    \centering
    \includegraphics[width=0.8\textwidth]{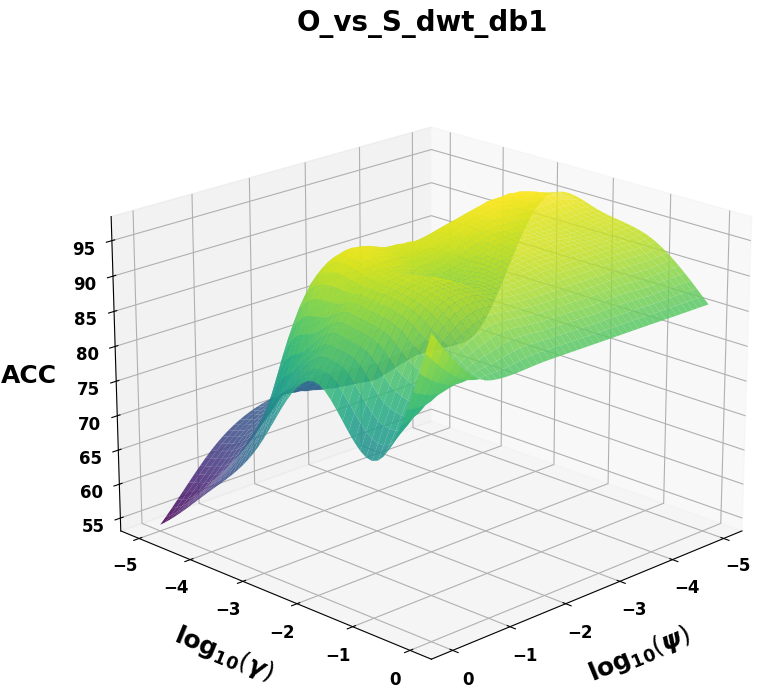}
  \end{minipage}\hfill
  \begin{minipage}[b]{0.45\textwidth}
    \centering
    \includegraphics[width=0.8\textwidth]{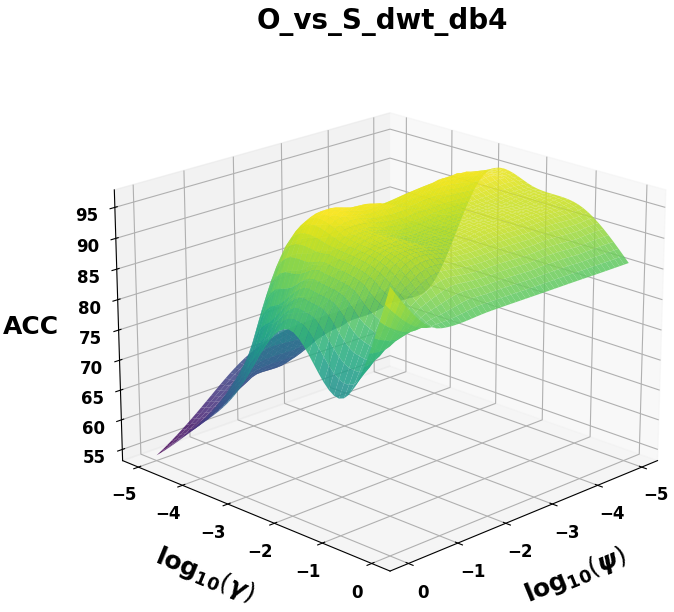}
  \end{minipage}\\[1ex]
  \begin{minipage}[b]{0.45\textwidth}
    \centering
    \includegraphics[width=0.8\textwidth]{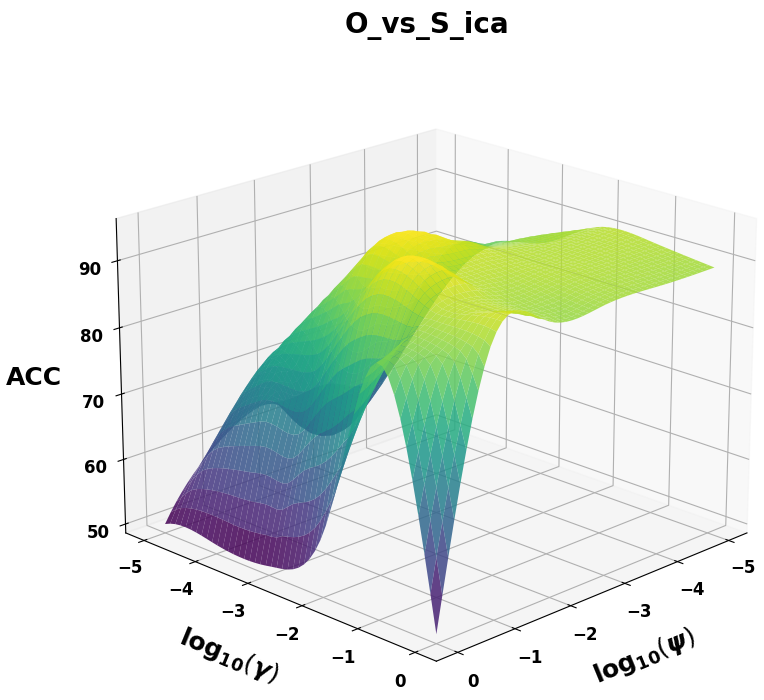}
  \end{minipage}\hfill
  \begin{minipage}[b]{0.45\textwidth}
    \centering
    \includegraphics[width=0.8\textwidth]{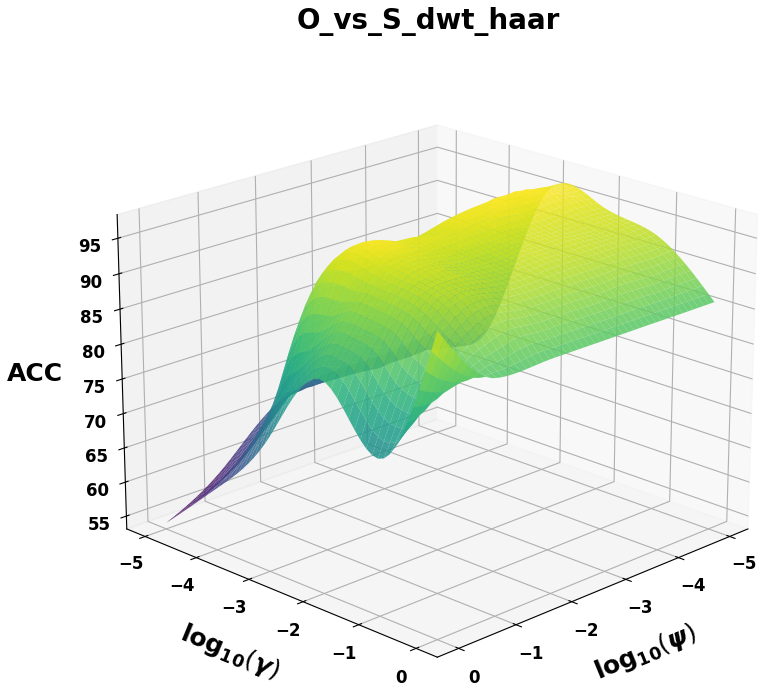}
  \end{minipage}
  \caption{Parameter sensitivity analysis showing the effect of $\gamma$ and $\psi$ on classification accuracy of IU-GEPSVM across different feature types.}
  \label{fig:AUC_c_rho}
\end{figure}

Based on the sensitivity analysis presented as 3D surface plots in Figure~\ref{fig:AUC_c_rho}, the IU-GEPSVM model exhibits distinct performance characteristics across different feature extraction methods. The analysis examines accuracy variation with respect to two logarithmic hyperparameters, $\log_{10}(\gamma)$ and $\log_{10}(\psi)$, for Haar, Daubechies-4, Daubechies-1 wavelets, and Independent Component Analysis.

DWT-based features demonstrate smooth, dome-like sensitivity surfaces peaking at 83--85\% accuracy in the central parameter region ($\log_{10}(\gamma) \approx -3$ to $0$, $\log_{10}(\psi) \approx -5$ to $-3$). The broad performance plateaus indicate robust classification across wide parameter ranges, with gradual degradation at extreme values. Critically, all three wavelet variants exhibit consistent behavior, suggesting model stability independent of the specific wavelet basis function.

Conversely, ICA features present a markedly different landscape with complex topology characterized by central valley structures and performance peaks (67--73\%) along parameter space boundaries. This narrower optimal region, coupled with substantially lower accuracy compared to DWT features (67\% versus 83--85\% for O\_vs\_S), indicates greater sensitivity to parameter misspecification. These contrasting sensitivity patterns reveal that wavelet-based features provide superior robustness and performance, requiring less stringent hyperparameter optimization, while ICA features necessitate more sophisticated tuning strategies for clinical deployment.

\section{Conclusion}

This study presented Universum-enhanced SVM frameworks aimed at improving EEG-based seizure classification by utilizing prior knowledge from interictal brain states. The proposed U-GEPSVM and IU-GEPSVM models demonstrated that incorporating Universum learning can effectively enhance class discrimination and robustness against EEG non-stationarity and limited data conditions. Among the two, IU-GEPSVM provided greater numerical stability and more reliable performance by reformulating the optimization into a standard eigenvalue problem.  

Comprehensive experiments on the Bonn University EEG dataset established consistent improvements in accuracy and statistical significance over conventional SVM-based methods. The integration of interictal Universum data proved beneficial in shaping more meaningful decision boundaries without the need for synthetic augmentation. While IU-GEPSVM introduced slightly higher computational cost, the performance gains justify its applicability in diagnostic scenarios where model reliability and interpretability are critical.  

Nonetheless, the approach assumes the availability of representative Universum data and was evaluated only for binary classification tasks. Future research will focus on extending the models to multi-class and multi-modal EEG analysis, investigating adaptive Universum selection strategies, and developing real-time or resource-efficient variants suitable for clinical and brain–computer interface systems.  

Overall, the findings suggest that Universum-based learning offers a promising direction for enhancing EEG classification reliability, providing a statistically validated and clinically relevant foundation for advancing computational tools in neurological diagnosis.

 \bibliographystyle{elsarticle-num} 
 \bibliography{references}
    \nocite{*}




\end{document}